\documentclass[10pt,twocolumn,letterpaper]{article}
\pdfoutput=1

\usepackage{iccv}
\usepackage{times}
\usepackage{epsfig}
\usepackage{graphicx}
\usepackage{amsmath}
\usepackage{amssymb}

\usepackage[utf8]{inputenc}
\usepackage[table,xcdraw]{xcolor}
\usepackage{colortbl}
\usepackage{enumitem}
\usepackage{soul}
\makeatletter
\@namedef{ver@everyshi.sty}{}
\makeatother
\usepackage{tikz}
\usepackage{gensymb}
\usepackage{booktabs}
\usepackage{multirow}
\usepackage{wrapfig}
\usepackage{capt-of}
\usepackage{authblk}
\usepackage[normalem]{ulem}
\useunder{\uline}{\ul}{}
\usepackage{subfig}

\usepackage[accsupp]{axessibility}
\usepackage[breaklinks=true,bookmarks=false]{hyperref}

\iccvfinalcopy % *** Uncomment this line for the final submission

\def\iccvPaperID{8} % *** Enter the ICCV Paper ID here

\ificcvfinal\fi

\newcommand{\norm}[1]{\lVert#1\rVert}

\begin{document}

%%%%%%%%% TITLE
\title{MonoCInIS: Camera Independent Monocular 3D Object Detection \\ using Instance Segmentation}

\author{First Author\\
Institution1\\
Institution1 address\\
{\tt\small firstauthor@i1.org}
\and
Second Author\\
Institution2\\
First line of institution2 address\\
{\tt\small secondauthor@i2.org}
}

\author{
    Jonas Heylen$^{1}$ \quad
    Mark De Wolf$^{1}$ \quad
    Bruno Dawagne$^{1}$ \quad
    Marc Proesmans$^{1,2}$ \quad
    Luc Van Gool$^{2,3}$ \hspace{0.1cm} \quad
    Wim Abbeloos$^{4}$ \quad
    Hazem Abdelkawy$^{4}$ \quad
    Daniel Olmeda Reino$^{4}$ \quad
    \vspace{3pt}\\
    $^1$TRACE vzw \quad $^2$KU Leuven/ESAT-PSI \quad $^3$ ETHZ/CVL \quad $^4$Toyota Motor Europe\\
    {\tt\small \{jonas.heylen,mark.dewolf,bruno.dawagne\}@trace.vision \{marc.proesmans,luc.vangool\}@esat.kuleuven.be \{Wim.Abbeloos,Hazem.Abdelkawy,Daniel.Olmeda.Reino\}@toyota-europe.com}
%    {\tt\small \{jonas.heylen,mark.dewolf,bruno.dawagne\}@trace.vision},
%    {\tt\small \{marc.proesmans,luc.vangool\}@esat.kuleuven.be}
%    {\tt\small \{Wim.Abbeloos,Hazem.Abdelkawy,Daniel.Olmeda.Reino\}@toyota-europe.com}\\
}

% \author[1]{Jonas Heylen \\ TRACE vzw}
% \author[1]{Mark De Wolf}
% \author[1]{Bruno Dawagne}
% \author[1,2]{Marc Proesmans}
% \author[2,3]{Luc Van Gool}
% \author[4]{Wim Abbeloos}
% \author[4]{Hazem Abdelkawy}
% \author[4]{Daniel Olmeda Reino}
% \affil[1]{Trace VZW}
% \affil[2]{KU Leuven} ESAT-PSI}
% \affil[3]{Toyota Motor Europe}

\maketitle
\ificcvfinal\thispagestyle{empty}\fi

%%%%%%%%% ABSTRACT

\label{abstract}
\begin{abstract}

Monocular 3D object detection has recently shown promising results, however there remain challenging problems. One of those is the lack of invariance to different camera intrinsic parameters, which can be observed across different 3D object datasets. Little effort has been made to exploit the combination of heterogeneous 3D object datasets. In contrast to general intuition, we show that more data does not automatically guarantee a better performance, but rather, methods need to have a degree of 'camera independence' in order to benefit from large and heterogeneous training data. In this paper we propose a category-level pose estimation method based on instance segmentation, using camera independent geometric reasoning to cope with the varying camera viewpoints and intrinsics of different datasets. Every pixel of an instance predicts the object dimensions, the 3D object reference points projected in 2D image space and, optionally, the local viewing angle. Camera intrinsics are only used outside of the learned network to lift the predicted 2D reference points to 3D. We surpass camera independent methods on the challenging KITTI3D benchmark and show the key benefits compared to camera dependent methods.

\end{abstract}

%%%%%%%%% BODY TEXT
\section{Introduction}
\label{section:introduction}

    % General introduction
    Predicting accurate 3D object position and orientation is crucial in the context of autonomous systems that interact with a set of objects in a common environment. A particularly relevant application is pose estimation of vehicles in autonomous driving. Where most of the initial efforts have been based on high precision LiDAR and stereo vision, simpler setups based on monocular vision %are desirable. Predicting 
    have gained interest. Nevertheless, 3D pose estimation  from monocular views remains a challenging task, as it is largely an ill-posed problem.
    \begin{figure}[t]
        \centering
        \includegraphics[width=0.8\linewidth]{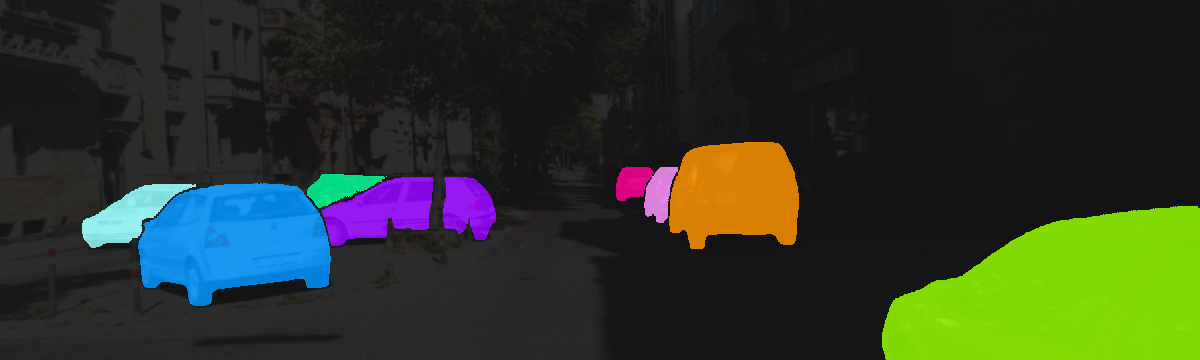}
        \includegraphics[width=0.8\linewidth]{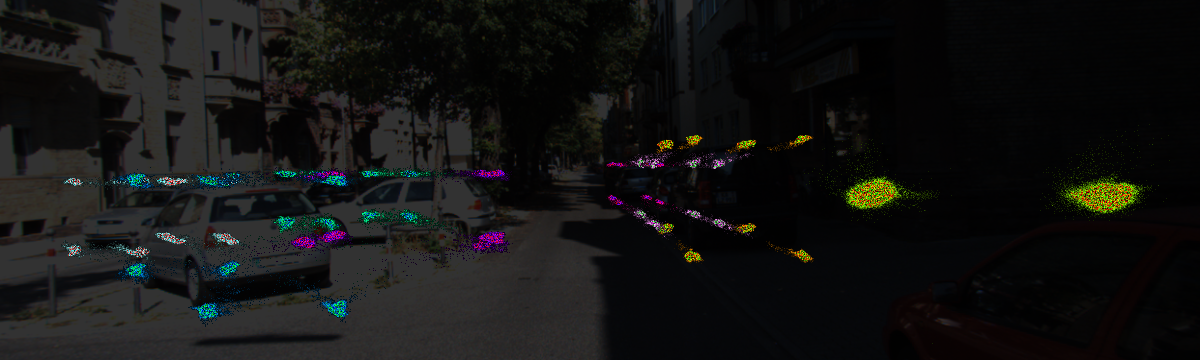}
        \includegraphics[width=0.8\linewidth]{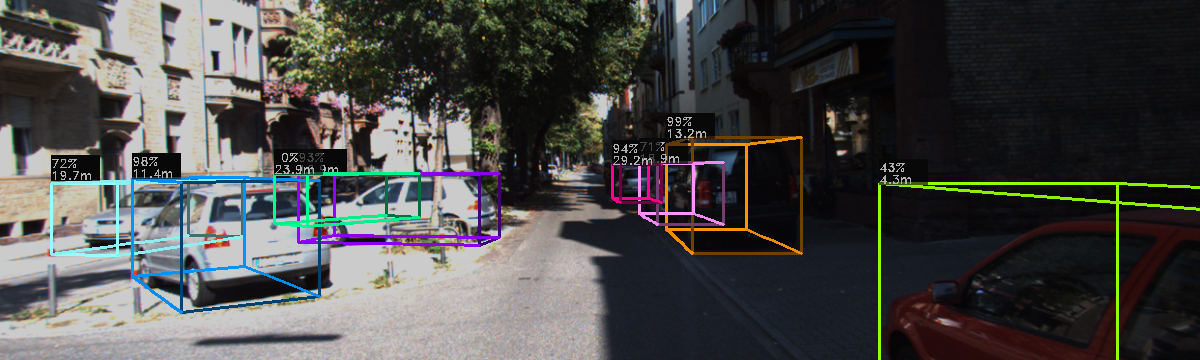}

        \caption{
        Our method leverages instance segmentation predictions (top), in which every pixel predicts the object's dimensions and projected 3D reference points (mid).
        % Proposal free instance segmentation (top), predicts the object's dimensions and projected 3D reference points (mid) for each pixel. 
        Camera independent geometric reasoning lifts these to 3D (bottom).
        %%%This approach is camera independent since camera intrinsics are only used during lifting and not learned within the network.
        }
        
        \label{fig:qual}
    \end{figure}
    %
    % New datasets available
    Recently, a number of large 3D detection datasets for autonomous driving have been made available. Starting with KITTI3D~\cite{Geiger2012CVPR}, others have followed, namely CityScapes3D~\cite{cordts2016cityscapes}, nuScenes~\cite{caesar2019nuscenes}, Waymo~\cite{sun2019scalability}, or Lyft~\cite{lyft2019}.
    % This gives new opportunities
    This provides an opportunity to exploit training data from heterogeneous sources,
    %It would entails though the need for methods 
    but it also suggest the need for methods which are able to handle this variety of cameras, with different intrinsics and viewing characteristics.
    %, and, not in the least, camera models which go beyond the ones that are available now.
    We refer to these methods as \textbf{camera independent}.
    % Extra explanation for camera independence 
    To the best of our knowledge, we are the first to investigate the effects of combining large datasets in the context of monocular 3D object detection for automated driving applications.

    % Why others fail
    Many state-of-the-art (SOTA) methods depend on regressing depth directly, either through estimating a full depth map or individual object distances. This results in the network learning an internal representation of the camera intrinsics, by linking position and scale in the image to real world depth. We show that, for such camera dependent methods, more data doesn't necessarily result in better performance. These methods lack the ability to handle an arbitrary number of views, but rather learn a few different camera models, while not being able to scale and generalise to any camera model. 
    % stress the fact that this is not only about the ability to cope with a changing focal length
    %% internal refs : \cite{yogamani2019woodscape,wiki-fe} \cite{Xie2016CVPR} \cite{wiki-pano}
    An extensive overview on camera models can be found in Sturm~\etal~\cite{sturm2011}.
    %fisheyes, catadioptric cameras, 
    % How to overcome this
    On the other hand, some methods rely solely on camera independent geometric reasoning, leveraging known camera intrinsics outside the learned network. However, existing camera independent methods lack performance.
    %
    
    % What we propose
    In this work, we propose a pose estimation method 
    which is able to take advantage of the high scene variability of multiple datasets. It is based on proposal free \textbf{instance segmentation} which avoids the need for Non-Maximum-Suppression (NMS).
    %
    % extract from [Distance-Normalized Unified Representation for Monocular 3D Object Detection] : An important step for monocular 3D object detection is Non-Maximum Suppression (NMS), which is usually based on the confidence from the classification branch [33, 1]. This may cause omissions of candidate boxes with high-quality3D information prediction, because the higher classification confidence doesn’t always interpret as the better 3D information prediction. 
    %
    Every pixel of an object's instance predicts multiple camera independent representation attributes, such as the object dimensions, the 3D reference points (RPs) projected into 2D image space and, optionally, a local viewing angle. Each pixel's vote contributes to the final prediction. 
    Camera intrinsics are used outside the learned network along with simple geometric reasoning to uplift the predicted 2D RPs to 3D. 
    We show that our method outperforms existing camera independent methods %by large margins 
    and achieves similar results as last year's camera dependent methods on the challenging KITTI3D benchmark. Recent works however surpass the performance of our method by strongly focusing on the KITTI3D dataset, but they lack the ability to generalise over heterogeneous cameras.
    We further investigate the trade-off between accuracy and speed, which is an important aspect for practical applications in the context of autonomous driving.
    % Maybe our contribution in bullet points?
    Our main contributions can be summarised as follows:
    \begin{itemize} [topsep=1pt,itemsep=1pt,partopsep=1pt, parsep=1pt]
    
    \item We propose a category-level 3D pose estimation method based on instance segmentation. 
    %which eliminates the need for NMS.
    Each pixel of an instance votes for all 
    %representation 
    attributes, resulting in distributions
    %, providing 
    with confidence estimates.
    \item To the best of our knowledge, we are the first to investigate the effects of combining different datasets in the context of monocular 3D object detection.
    \item We show qualitatively that our approach generalises well over different camera types such as fisheyes, even without fisheye training data.
    \item We release instance segmentation annotations for the KITTI3D dataset.
    \end{itemize}

\section{Related work}
\label{section:related work}

This section briefly reviews related works on 3D object detection using LiDAR data, stereo images, depth, 3D shape information and monocular images. 

\textbf{LiDAR data and stereo images.} 
In the field of autonomous driving, best results on 3D object detection challenges~\cite{Geiger2012CVPR, caesar2019nuscenes, sun2019scalability} are achieved by methods using LiDAR data~\cite{ye2020sarpnet, hestructure}, which can benefit from having reliable depth information. Stereo images can also provide depth information~\cite{chen2020dsgn, xu2020zoomnet}, to even mimic point cloud data based on RGB images only, leveraging the possibility to use existing LiDAR-based methods on so-called Pseudo-LiDAR point clouds~\cite{wang2019pseudo, you2019pseudo}.

\textbf{Monocular images.} Having no depth sensor data available, 3D object detection based on monocular images only is very challenging. Different strategies have been used to tackle this ill-posed problem. 
%
% (moved down) Predicting \textit{depth from monocular images} in a sub-network has been used to try to overcome this problem~\cite{xu2018multi, wang2019pseudo, you2019pseudo, wang2019task, ma2019accurate, weng2019monocular, ding2020learning, vianney2019refinedmpl, qian2020end}. One work uses Structure from Motion (SfM) cues together with 2D object detection to predict 3D bounding boxes in road scenes~\cite{song2015joint}.
%
One of the first approaches in monocular 3D object detection was introduced by DeepBox3D~\cite{mousavian20173d}, which solves 3D translation using \textit{geometric constraints} by predicting 3D orientation and dimensions for each 2D proposal. Several other works follow this approach and extend it in several ways, for example by visual cues~\cite{liu2019deep}, solving a closed form solution~\cite{naiden2019shift}, or integrating the 3D reconstruction into the network by reprojecting the predicted 3D box in both image space as Bird's Eye View (BEV)~\cite{min2019multi}. Also the use of segmentation masks leads to improved results for this approach~\cite{ku2019monocular, chen2016monocular}.
Other methods adopt a \textit{BEV} to predict bounding boxes~\cite{roddick2018orthographic, kim2019deep, srivastava2019learning}.
CaDDN\cite{reading2021cadnn} uses categorical depth distribution for each pixel to project contextual feature information to the appropriate depth interval in 3D space, and then uses a BEV projection to produce the final output bounding boxes. 

Another approach is to \textit{use a 2D detector}, and predict a 3D bounding box for each proposal. These works usually predict direct depth per detected 2D bounding box~\cite{ZhuJ2019objdist, zhou2019objects}. MonoGRNet~\cite{qin2019monogrnet} consists of four sub-networks: 2D detection, instance depth estimation, 3D location estimation and local corner regression. ROI-10D~\cite{manhardt2019roi} proposes a novel loss by lifting 2D detections, orientation and scale estimation into 3D space. MonoDIS~\cite{simonelli2019disentangling} proposes a two-stage method, disentangling dependencies of different parameters by introducing a novel loss enabling to handle them separately. SS3D~\cite{jorgensen2019monocular} and M3D-RPN~\cite{brazil2019m3d} are single-stage and feed the predictions to a 3D bounding box optimizer. MoVi-3D~\cite{simonelli2019single} generates virtual views where the object appearance is normalized with respect to distance from camera, reducing the visual appearance variability and relieving the model from learning depth-specific representations.
M3DSSD\cite{luo2021m3dssd} introduces feature alignment to avoid mismatching, and asymmetric non-local attention.
MonoDLE~\cite{ma2021monodle} investigates the misalignment between the center
of the 2D bounding box and the projected center of the 3D
object, and argues to remove distant objects since they mislead the network.
Other methods use \textit{3D Shape Information}~\cite{barabanau2019monocular}. Deep MANTA~\cite{chabot2017deep} consists of three levels of box refinement and predicts template similarity to known 3D shapes. 3D-RCNN~\cite{kundu20183d} learns a low-dimensional shape-space from a collection of 3D shape models. Mono3D++~\cite{he2019mono3d++} jointly optimises the 3D-2D consistency and task priors like monocular depth, ground plane projection and morphable shape model pose estimation.

In \cite{rangesh2018ground}, \textit{2D keypoints} of interest are predicted together with 2D bounding boxes, coarse orientations and dimensions. 3D boxes are reconstructed after polling predefined ground planes. RTM3D~\cite{li2020rtm3d} predicts 2D keypoints of the eight corners and 3D center as heatmaps. The 3D bounding box is computed using geometric constraints of the perspective projection. In \cite{cai2020monocular}, a structured polygon of 2D keypoints is predicted and projected in 3D using the object height prior. 
%All of the above works use predefined keypoints. 
KeypointNet~\cite{suwajanakorn2018discovery} describes a method to extract 3D keypoints from a single image using geometric reasoning. SMOKE~\cite{liu2020smoke} predicts a 3D bounding box for each detected object by combining single keypoint estimates with regressed 3D variables. In addition, MonoPair~\cite{chen2020monopair} predicts virtual pairwise constraint keypoints, the middle point of any two objects if they are the nearest neighbors. \cite{pavlakos20176} predicts keypoints using heatmaps.
LiteFPN~\cite{Yang2021LiteFPNFK} proposes a generic Lite-FPN module that conducts multi-scale feature fusion for their keypoint-based detectors.

In order to overcome the lack of real 3D information, several methods have included the prediction of \textit{depth from monocular images} in a sub-network~\cite{xu2018multi, ma2019accurate, weng2019monocular, vianney2019refinedmpl, qian2020end}. One work uses Structure from Motion (SfM) cues together with 2D object detection to predict 3D bounding boxes~\cite{song2015joint}. DDMP~\cite{Li2021depthmessprop} predicts depth-dependent filter weights and affinity matrices for information propagation.  D4LCN~\cite{ding2020learning} introduces depth-guided filters which are learned from image-based depth maps.  In \cite{Liu2021GroundAwareM3}, additional clues are identified from the ground plane and inserted in the depth reasoning to improve the positioning of the bounding boxes. MonoEF~\cite{zhou2021monoEF} provides a mechanism to cope with ego-car pose changes w.r.t. ground plane.

%(Paragraph related to uncertainty)
Some of the more recent methods include a form of \textit{uncertainty reasoning} to improve robustness.
MonoRUn~\cite{Chen2021MonoRUnM3} proposes a robust KL loss that minimizes the uncertainty-weighted reprojection error of the 3D coordinates onto the image plane.
%, and propagates it to an uncertainty-driven PnP algorithm to determine the pose.
MonoFlex\cite{zhang2021objdiff} uses edge fusion to decouple the feature learning and to predict truncated objects. Object depth is estimated using an uncertainty-guided ensemble of directly regressed depth and solved depths from different groups of keypoints.

%\textbf{Camera independence.} In this work, we define camera independence as not using depth or indirectly learning camera intrinsics from data. Within the monocular methods, several older methods meet this definition such as DeepBox3D~\cite{mousavian20173d}, Deep MANTA~\cite{chabot2017deep} and MonoGRNet2~\cite{barabanau2019monocular}. Most of them use geometric reasoning, prior shape information or 2D keypoints. Other methods could possibly also become camera independent by leaving out the camera dependent part of their system~\cite{naiden2019shift}. However, these methods mostly have poor performance, or don't report results.

\textbf{Segmentation based 6D pose estimation.} A number of works propose the use of semantic or instance segmentation as a means to estimate the 6D object pose.
PoseCNN~\cite{xiang2018posecnn} uses segmentation as detector, and predicts the object center by having each pixel of an instance vote for the direction. The distance is computed as the average of all pixels' votes. ConvPoseCNN~\cite{capellen2019convposecnn} extends the method by predicting also the rotation for every pixel. LieNet~\cite{do2018deep} proposes the decoupling of pose parameters so that the rotation can be regressed via Lie algebra representation. Several works use a similar approach to predict 2D keypoints, and use a PnP solution to compute the 3D pose~\cite{jafari2018ipose, li2019cdpn, zakharov2019dpod}. Optionally depth can be predicted as well~\cite{wang2019normalized}. Most of these works assume prior knowledge on object dimensions, shape or 3D cad models. They are designed to operate on specific objects, not on whole categories.
A broad overview on the current state-of-the-art on 6D pose detection and tracking is presented by Fan et al.~\cite{Fan2021DeepLO}. They cover the topic in the more general context, for applications ranging from autonomous driving, robotics and augmented reality.

\textbf{Camera independence.} In this work, we define camera independence as 'not using depth or indirectly learning camera intrinsics from data',
or alternatively, focusing on features that can be learned from appearance such as object size or 2D reference points.
Within the monocular methods, several older methods meet this definition such as DeepBox3D~\cite{mousavian20173d}, Deep MANTA~\cite{chabot2017deep} and MonoGRNet2~\cite{barabanau2019monocular}. 
Most of them use geometric reasoning, prior shape information or 2D keypoints. Other methods could possibly also become camera independent by leaving out the camera dependent part of their system~\cite{naiden2019shift}. However, these methods mostly have poor performance, or don't report results.

\section{Method}
\label{section:method}
    
%In this work, w
We propose a novel camera independent approach using instance segmentation as a detection mechanism. Section~\ref{section:camera_independence} covers the general concepts and geometric reasoning %we use 
to achieve camera independence. %while
Section~\ref{section:using_instances} illustrates how those concepts are applied in our instance-based approach.
    \begin{figure*}[t]
        \centering
        % \resizebox{0.48\textwidth}{!}{
        \resizebox{0.94\textwidth}{!}{
            \input{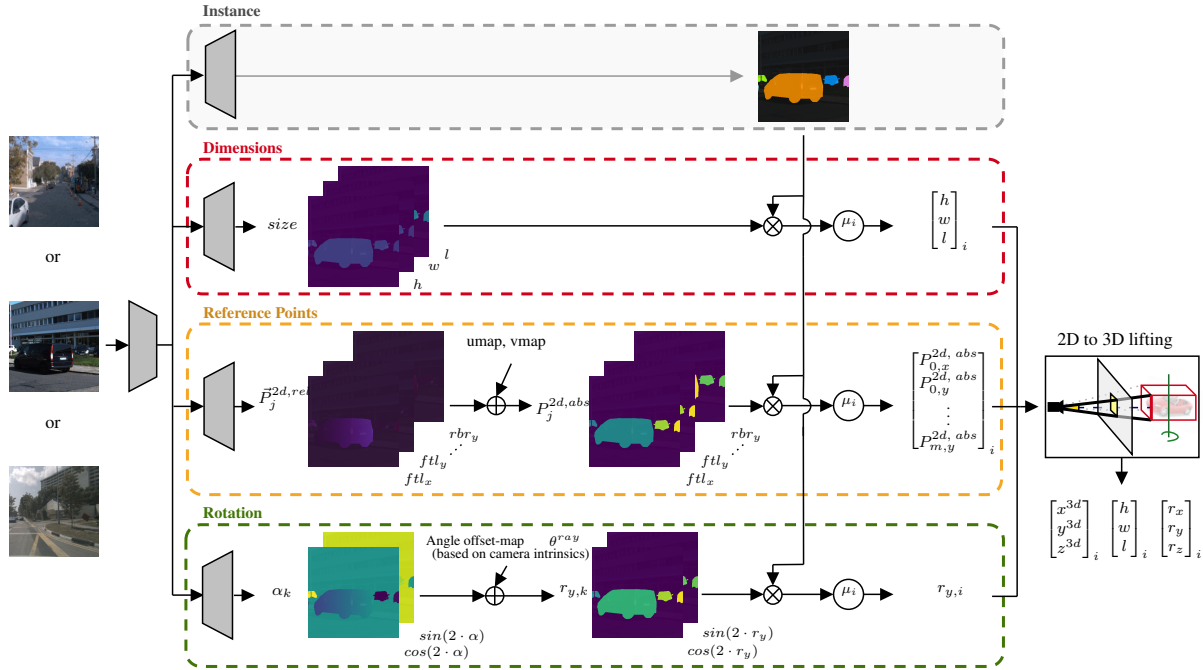}
        }
        \caption{Overview of our method. $\bigoplus$ denotes an addition: adding each pixel's coordinate to the predicted relative offset map or adding the angle offset for every pixel to the estimated viewing angles. $\bigotimes$ 
        %represents the process of 
        denotes masking out the predictions for each instance. \textcircled{$\mu_i$} represents the operation for averaging the predictions over all pixels belonging to each instance.
        }
        \label{fig:overview}
    \end{figure*}

    \subsection{Towards camera independence}
    \label{section:camera_independence}
    
    The main advantage of camera independence is the ability to combine training data from different datasets. In the scenario of autonomous vehicles, this means one network can handle a combination of multiple viewpoints and cameras, ranging from simple pinhole models to fisheye cameras
    %~\cite{wiki-fe} 
    and cylindrical projections~\cite{sturm2011}
    %~\cite{wiki-pano} 
    with different fields of view (FOV). Our method is able to generalize pose estimates for never-seen cameras (e.g. Fig.~\ref{fig:fisheye_qualitative}). This is useful for data where no ground truth is available during training, e.g. when a long-range camera view is out of LiDAR sensor range. 
    %This section discusses the different aspects of camera independence and substantiates our design choices.

\begin{figure}[t]
        \centering
        \resizebox{0.43\textwidth}{!}{
            \input{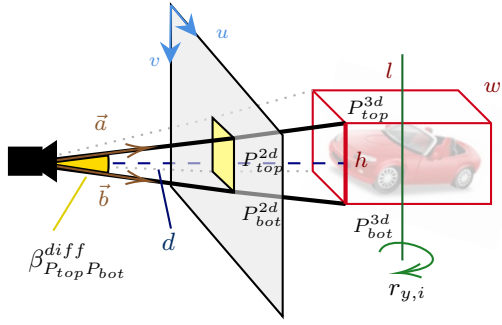}
        }
        \caption{Used notations and 2D to 3D lifting.}
        \label{fig:2Dto3D}
    \end{figure}

    \textbf{Object Dimensions.}
    As described in Section~\ref{section:related work}, convolutional neural networks (CNNs) are able to estimate object dimensions ($h$, $w$ and $l$ in Figure~\ref{fig:2Dto3D}) based on visual appearance. Training a network with images containing different viewpoints and crops, forces the network to become independent of camera-specific features or some internal representation of depth. Based on appearance only, the network can estimate the size of an object, regardless of the 
    perspective of the image.

    \textbf{2D Reference Points.}
    Most recent approaches directly predict the 3D distance for each object. Regressing this distance violates the above explained camera independence goal. We overcome this issue by predicting  \textit{Reference Points} (RPs) in the 2D image. These RPs are the 2D projections of predefined RPs in 3D, related to the object. A CNN can estimate the 2D RPs based on visual appearance rather than by learning a representation of the camera intrinsics, again contributing to the camera independence.
    This work explores two variants of predefined RPs. The first, \textit{8RP}, contains the 8 corners of the 3D bounding box surrounding the object. The second, \textit{2RP}, contains the top center and the bottom center of the 3D bounding box. Note that other combinations are possible.

    \begin{figure}[h]
        \centering
        \resizebox{0.32\textwidth}{!}{
             \input{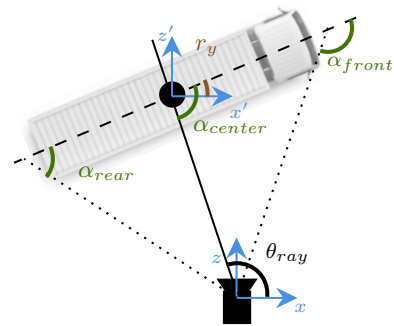}
        }
        \caption{Unique viewing angle (per pixel).}
        \label{fig:allocentric}
    \end{figure}
    
    \textbf{Object Rotation.}
    Multiple works have discussed the advantages of estimating the allocentric rather than the egocentric pose~\cite{manhardt2019roi, liu2019deep}. We predict the viewing angle, which constraints cars to be parallel to the ground-plane. Figure~\ref{fig:allocentric} shows the relationship between the \textit{viewing angle} $\alpha_{center}$ (allocentric) and the \textit{yaw} $r_y$ (egocentric) of an object. %
    As in~\cite{mousavian20173d}, estimating the sine and cosine avoids the discontinuity between $0\degree$ and $360\degree$.
    In some cases, the network is not able to distinguish the left and right side or the front and rear side of an object. When the network is confused between these $180\degree$ alternatives, it might predict the average viewing angle, resulting in a $90\degree$ offset.
    We disambiguate these cases by estimating the sine and cosine of $2\cdot\alpha$, resulting in a correct rotation without heading. To determine the correct heading, the sine and cosine of $\alpha$ could additionally be predicted and projected onto the $2\cdot\alpha$ vector.
    In this work, object rotation is required for the 2RP variant and optional for the 8RP variant.

    \textbf{Uplifting 2D to 3D.}
    Several existing methods can be used to map the 2D predicted RPs to a 3D object pose~\cite{jafari2018ipose, peng2019pvnet, hu2019segmentation, li2019cdpn, zakharov2019dpod}. We propose to calculate the 3D location of a pair of top-bottom RPs using simple trigonometry, as shown in Figure~\ref{fig:2Dto3D}. First, the angle $\beta^{diff}_{P_{top}P_{bot}}$ between the pixel rays $\vec{a}$, $\vec{b}$ is calculated for reference points $P^{2d}_{top}$ and $P^{2d}_{bot}$: $cos(\beta) = \vec{a}\cdot\vec{b}/(\norm{\vec{a}}\cdot\norm{\vec{b}})$ where $\vec{a}$ and $\vec{b}$ can be computed using camera intrinsics. As follows from Figure~\ref{fig:2Dto3D}, the distance $d$ between the camera and the center of $P^{2d}_{top}$ and $P^{2d}_{bot}$ is calculated as follows: $d = h/(2\cdot tan(\frac{\beta}{2}))$. Note that these equations 
    %are only valid when 
    assume the camera view is approximately perpendicular to the top-bottom line, which is a valid assumption in an autonomous vehicle scenario. Using the  distance $d$ and the pixel rays, the 2D RPs can be projected to 3D
    %One can easily generalise these equations where the perpendicular constraint is lifted and the full rotation of the object is used.
    providing an initial estimate of the 3D bounding box. In case of the 8RP variant, we can use the redundancy of the reference points to optimise the final 3D bounding box. This is done by updating the 3D box while minimising the distance between the 2D projections of the RPs in 3D, and the predicted 2D RPs. Similar to \cite{li2020rtm3d} we use the Levenberg–Marquardt algorithm (LM) to solve the optimisation problem. 
    %Note that camera intrinsics are needed during the uplifting. Since this is a post-processing step, the intrisincs are not embedded anywhere in the network.
    Note that camera intrinsics are needed during the uplifting, but since this is a post-processing step, the intrisincs are not embedded anywhere in the network.
    % so the camera independence is not violated.
    
    \subsection{Using instance segmentation}
    \label{section:using_instances}
    We propose to leverage instance segmentation masks to improve pose estimation. The key idea is that every pixel of an instance votes for each of the parameters described in Section~\ref{section:camera_independence}. 
    This results in vote distributions, which can be leveraged as a measure of confidence. This confidence can be used in a further stage of a broader pipeline, such as a tracking mechanism which requires confidence values.
    %(e.g. the AMOTA metric on the nuScenes benchmark).
    The use of instances has two other advantages. First, the prediction does not suffer from unreliable estimates caused by pixels which do not belong to the object itself, as is often the case for occluded objects in other approaches. Second, since our instance segmentation is proposal free, no %Non-Maximum-Suppression (NMS) 
    NMS is needed.
    Figure~\ref{fig:overview} shows an overview of our multi-task CNN. The encoder shares its weights for all tasks, while branched decoders have unique weights for each task. This section describes in more detail how the concepts of Section~\ref{section:camera_independence} are applied in this network.
    %and how they can benefit from an instance based detector.
    
    \textbf{Instance Segmentation.} The first branch outputs the instance segmentation. Note that any method can be used, even an external network or ground truth instances.
    
    \textbf{Object Dimensions.} The dimensions of an object are directly regressed for every pixel belonging to the instance mask of that object. Subsequently, all estimates belonging to that object's instance mask are averaged, as shown in the second branch of Figure~\ref{fig:overview}.
    
    \textbf{2D Reference Points.} Estimating the absolute coordinates for 2D RPs is not ideal in a CNN since the network has little knowledge of its absolute position within the image. 
    %Existing methods predict only the direction of the object center for each pixel~\cite{xiang2018posecnn}. 
    We propose to estimate full offset vectors between a pixel's coordinates $(u,v)$ and each 2D RP, for every pixel belonging to an object's instance mask $i$. Further on, we will call these offset vectors \textit{relative 2D RPs} $\vec{\textbf{P}}^{2d,rel}_{i,j}$, in contrast to the previously described \textit{absolute 2D RPs} $P^{2d,abs}_{i,j}$, where $j$ refers to a RP. Once the estimated relative RPs are converted to absolute RPs using Equation~\eqref{eq:rel2abs}, they are averaged over all pixels of each instance mask. This is shown in the third branch of Figure~\ref{fig:overview}. Note that the 2D positions of the absolute RPs are not limited by the image boundaries, as opposed to methods which predict heatmaps~\cite{pavlakos20176}.
    
    \begin{minipage}{0.45\textwidth}
        \begin{equation}
            \begin{aligned}
                P^{2d,abs}_{i,j_u,u,v} &= \vec{\textbf{P}}^{2d,rel}_{i,j_u,u,v} + umap_{u,v}\\
                P^{2d, abs}_{i,j_v,u,v} &= \vec{\textbf{P}}^{2d,rel}_{i,j_v,u,v} + vmap_{u,v}
            \end{aligned}
            \label{eq:rel2abs}
        \end{equation}
    \end{minipage}
    \begin{minipage}{0.1\textwidth}
        where
    \end{minipage}
    \begin{minipage}{0.3\textwidth}
        \begin{equation*}
            \begin{aligned}
                umap_{u,v} &= u\\
                vmap_{u,v} &= v
            \end{aligned}
        \end{equation*}
    \end{minipage}\\

    \textbf{Object Rotation.} Let us consider the example of the truck depicted in Figure~\ref{fig:allocentric}. The front part of the truck represents a considerably different viewing angle $\alpha_{front}$ compared to the rear part $\alpha_{rear}$. Based on this insight, we propose to predict a unique local viewing angle $\alpha$ for every pixel of an instance. As described in~\ref{section:camera_independence}, $cos(2\cdot\alpha)$ and $sin(2\cdot\alpha)$ are predicted. Every pixel's viewing angle is subsequently compensated by the pixel's light ray offset $\theta_{ray}$ to obtain the global yaw $r_y$. The instance masks are used to average the yaw for every instance, as shown in Figure~\ref{fig:overview}.

    \textbf{Uplifting 2D to 3D.}
    The dimensions, 2D RPs and rotation, can be combined to obtain full 6D object pose as described in Section~\ref{section:camera_independence} and shown in Figure~\ref{fig:2Dto3D}.

    % \done{\note{\jonas{Conclude by repeating advantages? Maybe not the right place to do it here? Would be more useful in conclusion?}
    %     - Issues with regular approach: NMS, some of these methods have difficulties with truncated cars, ... (Maybe find some references)\\
    %     - Advantages: occluded cars\\
    %     - Imagine the same car with pixels some pixels visible and occluded. Regular detector would predict on the occluded center the same value as not occluded. Instance based predicts the same offset values for visible pixels
    % }}

\section{Experiments}
\label{section:experiments}

\subsection{Setup}
\label{section:implementation}
\textbf{Our method.} This section discusses the implementation and training details for our method described in Section~\ref{section:method}. The method is implemented in PyTorch with off-the-shelf encoder-decoder architectures: ERFNet~\cite{romera2017erfnet}, ResNet50 and ResNet101~\cite{he2016deep}. Since the two ResNet backbones do not contain a decoder, we concatenate an ERFNet decoder to scale up the output to full image resolution. We initialise the ResNet backbones with pre-trained weights on ImageNet provided by PyTorch~\cite{torchvisionmodels}.
The first branch in Figure~\ref{fig:overview} implements the instance segmentation method proposed in \cite{neven2019instance}, since this method combines good accuracy with low inference time. All other branches are regressed with an $L1$ loss for pixels belonging to an instance, while background pixels are ignored. The rotation branch is only used for the $2RP$ variant. We use following weights on the losses: $[1, 45, 1]$ for the $8RP$ variant, and $[1, 40, 3, 10]$ for the $2RP$ variant.
We use NVIDIA~GTX~1080~Ti for training, evaluation and timing experiments. All used data is cropped and resized to a resolution of $720\times360$ or $1200\times360$.

\textbf{Other methods.} We compare our method with two other representative SOTA methods: SMOKE~\cite{liu2020smoke} and M3D-RPN~\cite{brazil2019m3d}. They both rank high on the KITTI3D benchmark and provide code to train on the KITTI3D dataset only~\cite{githubSMOKE, githubM3DRPN}, so we implemented additional dataloaders to be able to train on multiple datasets.

\begin{table}[ht]
\centering
\resizebox{0.46\textwidth}{!}{%
\begin{tabular}{@{}lllllllll@{}}
\toprule
\textbf{}                                         & \textbf{\begin{tabular}[c]{@{}l@{}}Images\\ (train+val)\end{tabular}} & \textbf{Cams}                                 & \textbf{Resolution}                                           & \textbf{FOV}                                           & \textbf{\begin{tabular}[c]{@{}l@{}}3D\\ boxes\end{tabular}} & \textbf{\begin{tabular}[c]{@{}l@{}}Inst.\\ Segm.\end{tabular}} & \textbf{Year} & \textbf{Locations}                                          \\ \midrule
\rowcolor[HTML]{EFEFEF} 
\textbf{KITTI3D}            & 3712+3769$^{1}$                                                       & 1                                             & 1242*375$^{4}$                                                & $81º$                                                  & Yes                                                         & Yes$^{2}$                                                      & 2012          & Germany                                                     \\
\textbf{VirtKITTI}       & 21260                                                                 & 1                                             & 1242*375                                                      & $81º$                                                  & Yes                                                         & Yes                                                            & 2016          & Germany                                                     \\
\rowcolor[HTML]{EFEFEF} 
\textbf{CityScapes3D} & 2975+500                                                              & 1                                             & 2048*1024                                                     & $48º$                                                  & Yes                                                         & Yes                                                            & 2016          & Germany                                                     \\
\textbf{SemKITTI}          & 200                                                                   & 1                                             & 1242*375                                                      & $81º$                                                  & No                                                          & Yes                                                            & 2018          & Germany                                                     \\
\rowcolor[HTML]{EFEFEF} 
\textbf{nuScenes}       & 168780+36114                                                          & \begin{tabular}[c]{@{}l@{}}5\\ 1\end{tabular} & 1600*900                                                      & \begin{tabular}[c]{@{}l@{}}$70º$\\ $110º$\end{tabular} & Yes                                                         & Yes$^{3}$                                                      & 2019          & \begin{tabular}[c]{@{}l@{}}Boston,\\ Singapore\end{tabular} \\
\textbf{Waymo}          & \begin{tabular}[c]{@{}l@{}}77499+13676\\ 54196+9564\end{tabular}      & \begin{tabular}[c]{@{}l@{}}3\\ 2\end{tabular} & \begin{tabular}[c]{@{}l@{}}1920*1280\\ 1920*1040\end{tabular} & $50º$                                                  & Yes                                                         & Yes$^{3}$                                                      & 2019          & USA                                                         \\
\rowcolor[HTML]{EFEFEF} 
\textbf{Lyft}                     & 136080                                                                & 6                                             & \begin{tabular}[c]{@{}l@{}}1224*1024\\ 1920*1080\end{tabular} & \begin{tabular}[c]{@{}l@{}}$70º$\\ $82º$\end{tabular}  & Yes                                                         & Yes$^{3}$                                                      & 2019          & Palo Alto                                                   \\ \bottomrule
\end{tabular}

}
\caption{Overview of used datasets. $^{1}$Chen split~\cite{chen2016monocular}, $^{2}$Manually annotated, $^{3}$Generated with SOTA method~\cite{massa2018mrcnn}, $^{4}$Resolution slightly varies.}
\label{tab:datasets}
\end{table}

\subsection{Datasets and Benchmarks}
\label{section:datasets}
In the field of autonomous driving, several datasets and benchmarks exist. Table~\ref{tab:datasets} summarises the datasets we use in our experiments.
\textbf{KITTI3D}~\cite{Geiger2012CVPR} is arguably the most popular benchmark for monocular 3D object detection. We follow the train-validation split described in~\cite{chen2016monocular}. We report on \textit{Car} only, both on the validation set and on the test set, using the official 3D object detection and BEV Average Precision (AP) metrics. The metric uses an Intersection-over-Union (IoU) of $0.7$ for \textit{Car}, which makes it especially hard for monocular object detection methods. Following~\cite{simonelli2019disentangling}, we report using the $AP_{3D,BEV|R_{40}}$ metrics, unless stated otherwise.
\textbf{VirtKITTI}~\cite{gaidon2016virtual} was created by virtually cloning real-world video sequences from KITTI3D, and providing automatic dense labeling.
\textbf{CityScapes}~\cite{cordts2016cityscapes} and \textbf{SemKITTI}~\cite{Alhaija2018IJCV} both provide instance segmentation annotations. The recently launched CityScapes3D dataset adds 3D object annotations.
\textbf{nuScenes}~\cite{caesar2019nuscenes}, \textbf{Waymo}~\cite{sun2019scalability} and \textbf{Lyft}~\cite{lyft2019} are recent big-scale 3D object detection datasets, captured by five or six cameras mounted on top of the car. Both nuScenes and Lyft cover the complete $360º$ surround view. We will refer to these three datasets as \textbf{NWL}.

\textbf{Generation of missing instance annotations.} Our method requires instance segmentation annotations. CityScapes, SemKITTI and VirtKITTI provide these annotations. For KITTI3D, we manually annotated all images with instance segmentation and will release them publicly. Since NWL do not provide these annotations, we generated pseudo-ground truth instances using a SOTA instance segmentation method~\cite{massa2018mrcnn}. %The generated instance masks are subsequently matched to the 3D bounding box annotations with the highest overlap in the image, starting with the closest bounding boxes. Note that this step is not part of our method, it only serves to provide the required annotations.

\begin{figure}[h]
	\centering
    \includegraphics[trim={0.5cm 0.6cm 0.3cm 1.1cm},clip,width=0.6\linewidth]{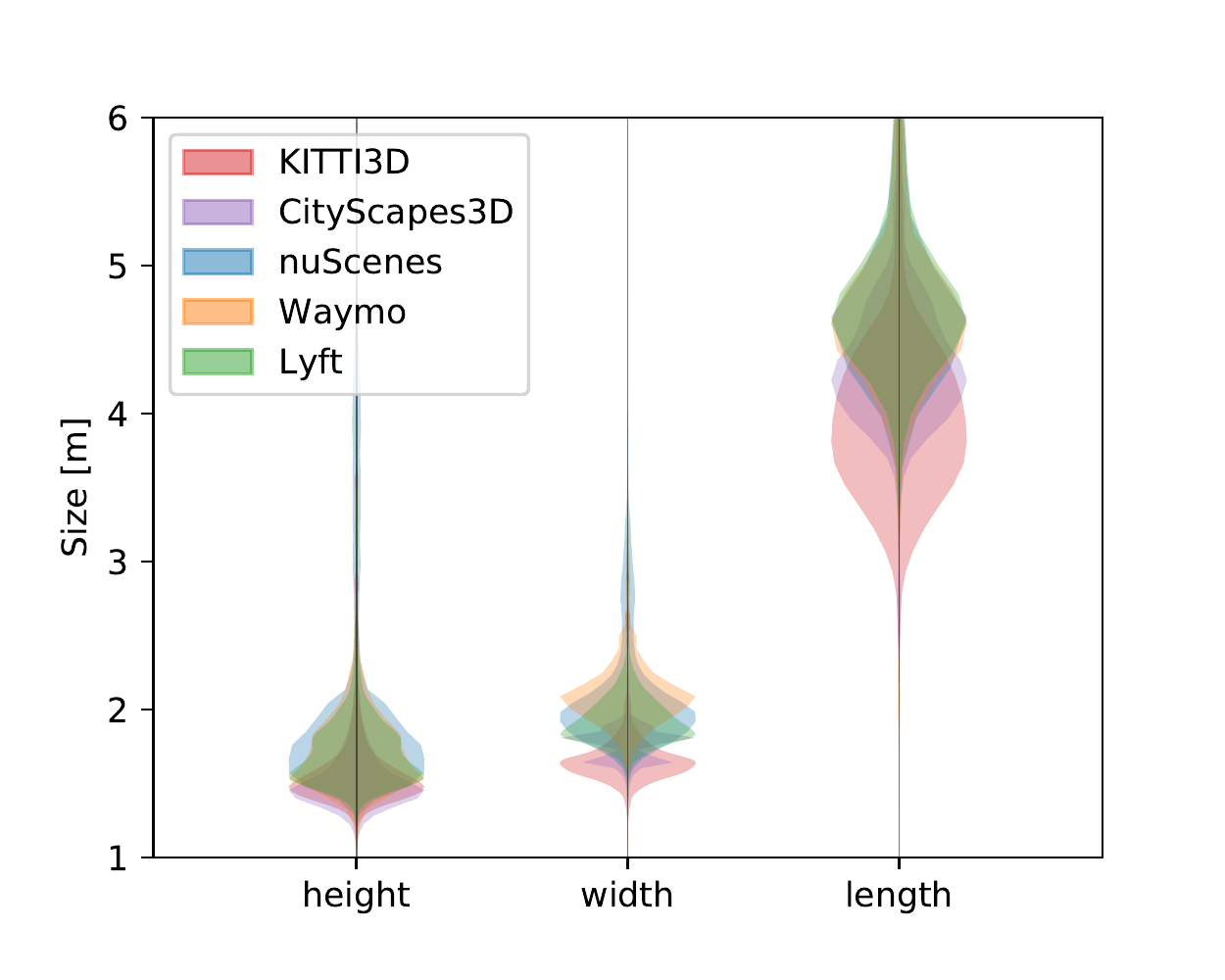}
    \caption{Imbalance in annotated car dimensions: KITTI3D and CityScapes3D overlap, while NWL is very different.}
    \label{fig:datasets-violin}
\end{figure}

\textbf{Annotation differences.} We address two observations in comparing different datasets. First, 3D bounding boxes 
%are annotated 
annotations are very tight in KITTI3D and CityScapes3D, while they are wider in NWL. Second, there is an imbalance in car size between the datasets, as shown in Figure~\ref{fig:datasets-violin} and described earlier in a LiDAR context~\cite{wang2020train}. KITTI3D and CityScapes3D have semi-overlapping distributions, while NWL is quite different.
%These observations encouraged us to consider the training with the required amount of care. 

%\sub
\section{Results and Discussion}
\label{section:results}

\begin{table}[h]
\centering
\resizebox{0.48\textwidth}{!}{%
\begin{tabular}{@{}lcl|cccccc@{}}
\toprule

                                                                                               &    &           & \multicolumn{6}{c}{Evaluation set results}                                                                              \\
                                                                                               &    & Training  & \multicolumn{3}{c|}{AP$_{3D|R_{40}}$}                                  & \multicolumn{3}{c}{AP$_{BEV|R_{40}}$}              \\
Method                                                                                         &    & datasets  & Easy           & Moderate       & \multicolumn{1}{c|}{Hard}          & Easy           & Moderate       & Hard           \\ \midrule
\multirow{4}{*}{\begin{tabular}[c]{@{}l@{}}Ours (8RP)\\ CI: Yes\end{tabular}}                  & BL & K3D       & 12.51          & 7.53           & \multicolumn{1}{c|}{6.21}          & 18.34          & 11.22          & 9.34           \\
                                                                                               & E1 & CS3D      & 2.78           & 1.13           & \multicolumn{1}{c|}{0.82}          & 6.21           & 3.21           & 2.55           \\
                                                                                               & E2 & K3D, CS3D & \textbf{16.16} & 8.80           & \multicolumn{1}{c|}{7.43}          & \textbf{23.14} & 12.78          & 10.98          \\
                                                                                               & E3 & K3D, NWL  & 16.09          & \textbf{9.19}  & \multicolumn{1}{c|}{\textbf{7.90}} & 22.90          & \textbf{13.86} & \textbf{11.53} \\ \midrule
\multirow{4}{*}{\begin{tabular}[c]{@{}l@{}}SMOKE~\cite{liu2020smoke}\\ CI: No\end{tabular}}    & BL & K3D       & \textbf{6.97}  & \textbf{4.37}  & \multicolumn{1}{c|}{\textbf{3.96}} & \textbf{12.01} & \textbf{8.03}  & \textbf{6.94}  \\
                                                                                               & E1 & CS3D      & 0.00           & 0.00           & \multicolumn{1}{c|}{0.00}          & 0.00           & 0.00           & 0.00           \\
                                                                                               & E2 & K3D, CS3D & 4.81           & 3.80           & \multicolumn{1}{c|}{3.03}          & 9.24           & 7.10           & 6.05           \\
                                                                                               & E3 & K3D, NWL  & 0.24           & 0.19           & \multicolumn{1}{c|}{0.19}          & 0.98           & 0.76           & 0.70           \\ \midrule
\multirow{4}{*}{\begin{tabular}[c]{@{}l@{}}M3D-RPN~\cite{brazil2019m3d}\\ CI: No\end{tabular}} & BL & K3D       & 14.68          & \textbf{10.76} & \multicolumn{1}{c|}{8.28}          & 21.52          & 15.59          & 12.44          \\
                                                                                               & E1 & CS3D      & 0.07           & 0.03           & \multicolumn{1}{c|}{0.03}          & 0.27           & 0.18           & 0.17           \\
                                                                                               & E2 & K3D, CS3D & \textbf{14.95} & 10.71          & \multicolumn{1}{c|}{\textbf{8.50}} & \textbf{22.55} & \textbf{16.34} & \textbf{12.79} \\
                                                                                               & E3 & K3D, NWL  & -              & -              & \multicolumn{1}{c|}{-}             & -              & -              & -              \\ \bottomrule
\end{tabular}%
}
\caption{Results on \textit{Car} (0.7 IoU) for the KITTI3D evaluation set. \textbf{Bold} refers to best performance within each method. CI: Camera Independent, BL: baseline, E: experiment, K3D: KITTI3D (train), CS3D: CityScapes3D, NWL: nuScenes (train), Waymo and Lyft. Note that the $R_{40}$ variant of AP is used.}
\label{tab:results-moredata}
\end{table}

In this section we discuss some results and insights of our experiments. 
%Each paragraph clearly mentions which datasets are used to train and evaluate, referring to section~\ref{section:datasets}. 
Unless stated otherwise, all experiments on our method start from a model pre-trained for instance segmentation on KITTI3D (train), VirtKITTI, CityScapes (train), SemKITTI (train) and NWL (see Section~\ref{section:datasets}).

\textbf{Multiple datasets and other methods.}
We verify two hypotheses on combining multiple datasets, comparing our camera independent method with two camera dependent SOTA methods: SMOKE~\cite{liu2020smoke} and M3D-RPN~\cite{brazil2019m3d}.
Table~\ref{tab:results-moredata} summarises the results of experiments \textbf{E1}-\textbf{E3} on the KITTI3D evaluation set, in comparison with the baseline (BL), trained on KITTI3D only.

\noindent \textit{\textbf{Hypothesis 1}: Training on dataset A and evaluating on dataset B works for camera independent methods, while camera dependent methods will fail to provide correct depth estimations when the FOV's of A and B differ substantially.}
%when A and B have very different FOV.}

    \setlength{\leftskip}{0.3cm}\noindent\textbf{E1}: trained on CityScapes3D only: our method gives reasonable results, especially since KITTI3D is never seen during training. Both SMOKE and M3D-RPN fail to provide accurate depth predictions.

\setlength{\leftskip}{0pt}\noindent\textit{\textbf{Hypothesis 2}: Camera dependent methods could possibly learn a few different camera models, but are not able to scale and generalise to any different camera model.}

    \setlength{\leftskip}{0.3cm}\noindent\textbf{E2}: trained on joint KITTI3D, CityScapes3D: all methods can cope, however SMOKE already deteriorates.

    \setlength{\leftskip}{0.3cm}\noindent\textbf{E3}: trained on joint KITTI3D, NWL: while SMOKE fails, our method can handle the wide variety of viewpoints, and even benefits from it. Note that the M3D-RPN implementation requires the \textit{occluded} and \textit{truncated} tags, which are not provided by NWL. This results in non-convergent trainings.

\setlength{\leftskip}{0pt}

\noindent For our method, the ResNet101 backbone and $8RP$ variant are used. For both SMOKE and M3D-RPN, we use the provided backbone in their implementations~\cite{githubSMOKE, githubM3DRPN}. Note that we should only compare within a method, since the training schemes are not optimised between methods. For SMOKE, we used a training scheme different from the original paper which explains the lower results. Nevertheless, the trainings with more data follow a longer training scheme, and thus are comparable in a fair way.

% Taking the trainings on KITTI3D (train) in each first row as a baseline, we describe some observations. First, our method achieves the best results when training on CityScapes3D only. Second, when adding other datasets to the training, our method is the only one which consistently improves. Not only when adding a similar dataset like CityScapes3D, but also with quite different datasets like NWL. Both observations prove the camera independent nature of our method, which can handle a wide variety of viewpoints and conditions, and can actually benefit from it without additional efforts.

% Overview of cam independent:
% Deepbox3D (or Deep3dBox?) 
    %   - https://arxiv.org/pdf/1612.00496.pdf
% Monocular 3D OD via Gemetric Reasoning on KP
    %   - https://arxiv.org/pdf/1905.05618.pdf
    %   - TODO: add
% Shift RCNN 
    %   - https://arxiv.org/pdf/1905.09970.pdf
    %   - This method's fist stage is independent, (I think they call it Lenear System in their results) Would be nice to mention both (in two rows) to be able to compare
    %   - TODO: add
% RTM3D
    %   - https://arxiv.org/pdf/2001.03343.pdf
    %   - This method also has an independent (I think) variant in the ablation study which actually outperforms ours...
%
% Camera dependent
% Mono 3D
%   - https://arxiv.org/pdf/2001.03343.pdf

% SOURCES evaluation set results:
% - https://arxiv.org/pdf/1905.05618.pdf
% - https://arxiv.org/pdf/2001.03343.pdf
% - https://arxiv.org/pdf/2002.10111.pdf
% - https://arxiv.org/pdf/1905.09970.pdf
% - https://arxiv.org/pdf/1912.04799.pdf
% - https://arxiv.org/pdf/2002.01619.pdf
% - https://arxiv.org/pdf/2003.00504.pdf

\textbf{Comparison to Related Work.} Table~\ref{tab:results-relatedwork} reports results on the KITTI3D benchmark on \textit{Car}, both on the test set and the evaluation set. For this experiment, we use the ResNet101 backbone. We train only on KITTI3D for both instance segmentation pre-training and 3D object detection, with no additional datasets. For the test set results, we use the full dataset. For the evaluation set results, we use only the train split.
%
% The $2RP$ variant outperforms the $8RP$ variant both on the 3D detection and BEV benchmark on the test set results. 
We compare only to methods which use RGB, without any additional LiDAR or depth data.
We outperform %all 
camera independent methods significantly. %with large margins.
When comparing with camera dependent methods, which over-fit to the specific camera intrinsics of the KITI3D dataset, we perform similar to last year's works. Recent methods however surpass our performance, at the cost of lacking the ability to generalise well over multiple cameras. \\

%%
%Second, compared to camera dependent methods, we outperform methods using BEV approaches like OFTNet~\cite{roddick2018orthographic} or directly predict depth like MonoGRNet~\cite{qin2019monogrnet}. We score similar or lower compared to more complex approaches like MonoDIS~\cite{simonelli2019disentangling}, SS3D~\cite{jorgensen2019monocular}, M3D-RPN~\cite{brazil2019m3d} and MoVi-3D~\cite{simonelli2019single}. Other recent methods which use a 2D keypoint approach like SMOKE~\cite{liu2020smoke} and RTM3D~\cite{li2020rtm3d} get similar results. 

%Note that SS3D, M3D-RPN and MoVi-3D are capable of predicting multiple classes.

% \begin{figure*}
%\noindent\begin{minipage}[h]{0.475\textwidth}
\noindent\begin{minipage}[h]{0.5\textwidth}
    \centering
    \resizebox{0.95\textwidth}{!}{%
    \begin{tabular}{@{}ll|ccc|ccc|cc@{}}
    \toprule
                               &     & \multicolumn{6}{c|}{Evaluation set results}                                                       & \multicolumn{2}{c}{\multirow{2}{*}{\begin{tabular}[c]{@{}c@{}}Inference\\ time [ms]\end{tabular}}} \\
                               &     & \multicolumn{3}{c|}{AP$_{3D|R_{40}}$}            & \multicolumn{3}{c|}{AP$_{BEV|R_{40}}$}             & \multicolumn{2}{c}{}                                                                               \\
                               &     & Easy           & Moderate      & Hard          & Easy           & Moderate       & Hard           & PyT                                              & TRT                                             \\ \midrule
    \multirow{2}{*}{ERFNet}    & 8RP & 6.34           & 4.14          & 3.22          & 10.79          & 6.96           & 5.63           & \textbf{29}                                      & \textbf{19}                                     \\
                               & 2RP & 5.26           & 3.19          & 2.61          & 8.21           & 5.04           & 4.02           & 34                                               & 21                                              \\ \midrule
    \multirow{2}{*}{ResNet50}  & 8RP & 14.66          & 7.71          & 6.63          & 20.73          & 11.19          & 9.66           & 101                                              & 89                                              \\
                               & 2RP & 14.31          & 8.08          & 6.46          & 20.01          & 11.68          & 10.20          & 112                                              & 103                                             \\ \midrule
    \multirow{2}{*}{ResNet101} & 8RP & 12.51          & 7.53          & 6.21          & 18.34          & 11.22          & 9.34           & 151                                              & 129                                             \\
                               & 2RP & \textbf{17.22} & \textbf{9.36} & \textbf{7.43} & \textbf{22.84} & \textbf{13.03} & \textbf{11.29} & 159                                              & 140                                             \\ \bottomrule
    
    \end{tabular}%
    }
    \captionof{table}{Results on \textit{Car} (0.7 IoU) for the KITTI3D evaluation set. \textbf{Bold} refers to best performance. PyT: PyTorch, TRT: TensorRT. Note that the $R_{40}$ variant of AP is used.}
    \label{tab:results-speedaccuracy}
\end{minipage}
    % \hspace{0.05\textwidth}
\begin{figure}[h]
	\centering
    \includegraphics[trim={0.5cm 0.4cm 0.3cm 0.3cm},clip,width=0.3\textwidth]{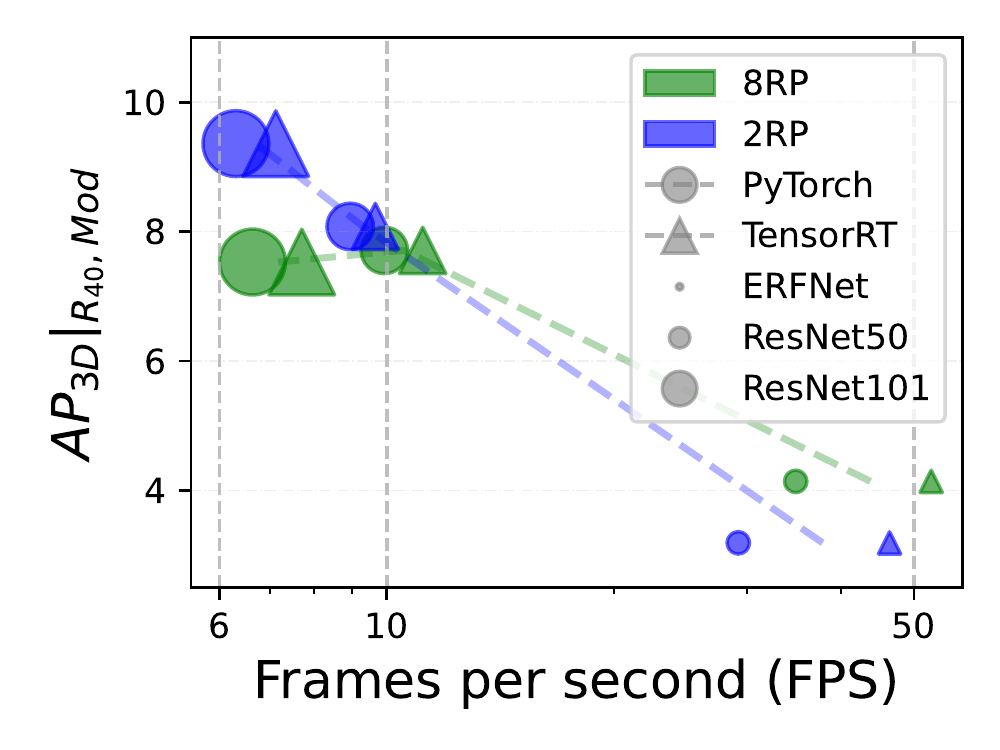}
    \captionof{figure}{Speed-accuracy trade-off: ResNet achieves higher accuracy, while ERFNet is faster.}
    \label{fig:results-timings}
\end{figure}

\textbf{Inference time.} Table~\ref{tab:results-speedaccuracy} compares results on the KITTI3D evaluation set for all three architectures described in Section~\ref{section:implementation}. We use only KITTI3D (train) during training. As expected, smaller network architectures achieve lower performance. However, the inference time also decreases. This leads to higher Frames per second (FPS). We report inference time using both PyTorch and TensorRT~\cite{tensorrt}. Figure~\ref{fig:results-timings} shows the trade-off between speed and accuracy.

\begin{figure}[b]
    \centering
    \subfloat[8RP model trained on KITTI3D only]{
        \label{kitti_fisheye}
        \begin{tabular}[b]{c}
            % % Old calib
            % \includegraphics[width=0.9\linewidth]{Figures/Results_qualitative_fisheye/8RP_train_kitti_eval_fisheye_boxes3d_pred.png}\\
            % \includegraphics[width=0.9\linewidth]{Figures/Results_qualitative_fisheye/8RP_train_kitti_eval_fisheye_boxes3d_pred_BEV.png}\\
            % New calib marc
            \includegraphics[width=0.9\linewidth]{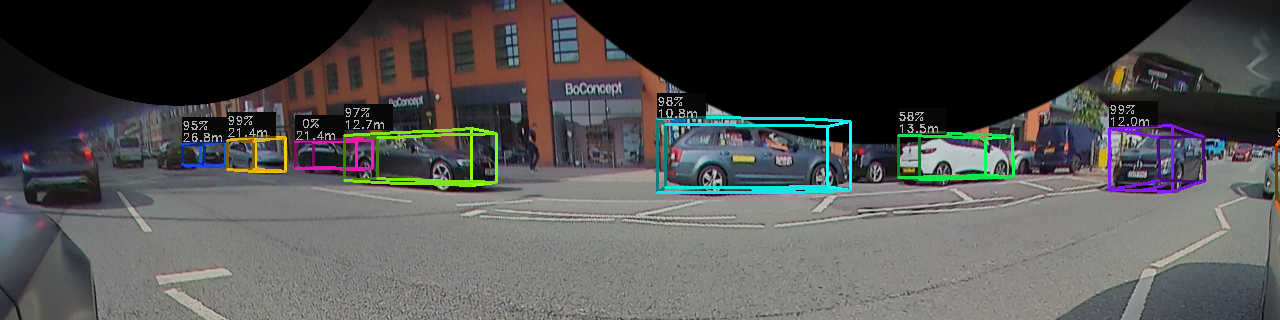}\\
            \includegraphics[width=0.9\linewidth]{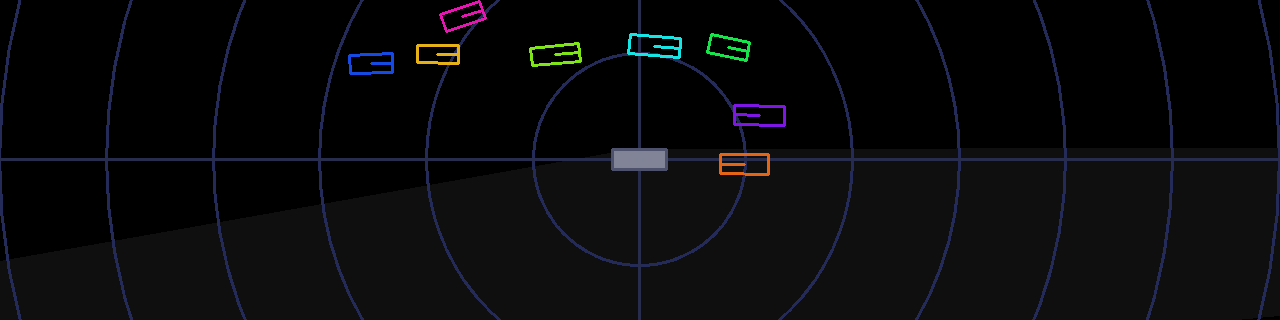}
        \end{tabular}
    }\\
    \subfloat[8RP model trained on KITTI3D and NWL]{
        \label{kitti_NWL_fisheye}
        \begin{tabular}[b]{c}
            % % Old calib
            % \includegraphics[width=0.9\linewidth]{Figures/Results_qualitative_fisheye/8RP_train_kitti+NWL_eval_fisheye_boxes3d_pred.png}\\
            % \includegraphics[width=0.9\linewidth]{Figures/Results_qualitative_fisheye/8RP_train_kitti+NWL_eval_fisheye_boxes3d_pred_BEV.png}\\
            % New calib marc
            \includegraphics[width=0.9\linewidth]{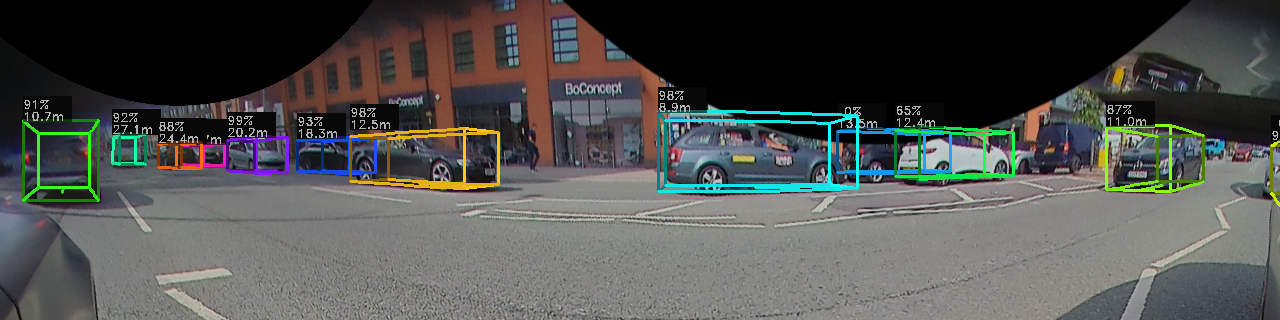}\\
            \includegraphics[width=0.9\linewidth]{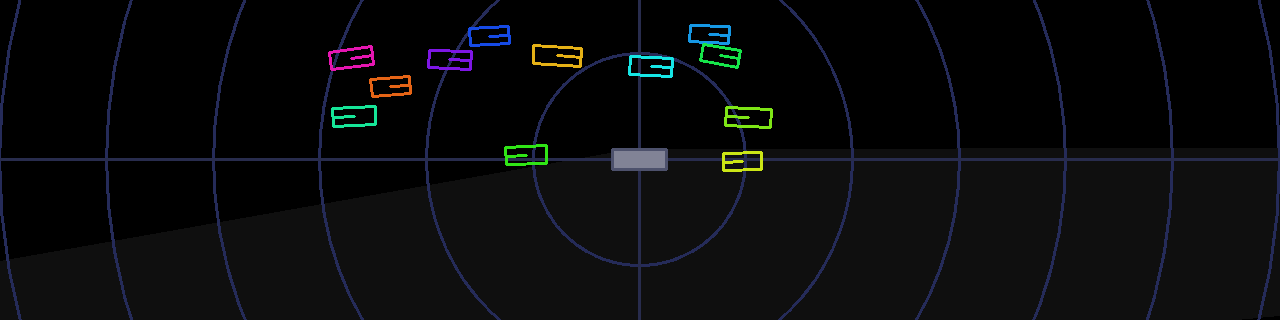}
        \end{tabular}
    }   
    \caption{Qualitative results %show that our method generalises well
    on unseen camera views: our model predicts 3D boxes on equirectangular fisheye images from ~\cite{yogamani2019woodscape}, though it was never trained for it. (a) vs. (b): generalisation improves when training on more varied data.
    }
    \label{fig:fisheye_qualitative}
\end{figure}

\textbf{Observations.} First, our method is reasonably robust to occlusions. In the presence of comparatively large occluders, the predicted RPs tend to shift towards them. Second, our method can be sensitive to small variations in 2D RPs or dimensions, which possibly lead to larger variations in 3D at far distance. 
Third, experiments on NWL show that our method is able to generalise easily to viewpoints within the scope of the training data. Although it is not meant to extrapolate to viewpoints that are out of the scope of the available datasets, 
Figure~\ref{fig:fisheye_qualitative} shows qualitatively that our method even copes with 
%unseen fisheye images. Note that we were not able to train on WoodScape~\cite{yogamani2019woodscape} data, since no 3D bounding box annotations are released yet.
unseen fisheye images from WoodScape~\cite{yogamani2019woodscape}. Note that we're not able to train on the data since no 3D bounding box annotations are released yet.
Figure~\ref{fig:qual} shows qualitative results of the $8RP$ variant of our method on a sample from the KITTI3D test set. 
%From top to bottom, it shows the instance segmentation, the 2D RPs, and the 3D bounding boxes. 
The proposed method is able to predict correct 3D bounding boxes under heavy occlusion and can recover from imperfect instance masks. More examples are provided in the supplementary material.

\begin{table*}[ht!]
\centering
\resizebox{0.99\textwidth}{!}{%
\begin{tabular}{@{}lcc|ccc|ccc|ccc|ccc@{}}
\toprule
                                               &      &              & \multicolumn{6}{c|}{Test set results}                                                                                                                                                & \multicolumn{6}{c}{Evaluation set results}                                                                                                                                                                                           \\ \midrule
                                               &      & Camera       & \multicolumn{3}{c|}{AP$_{3D|R_{40}}$}                                                      & \multicolumn{3}{c|}{AP$_{BEV|R_{40}}$}                                                      & \multicolumn{3}{c|}{AP$_{3D|R_{11}}$}                                                                                                     & \multicolumn{3}{c}{AP$_{BEV|R_{11}}$}                                                        \\
Method                                         & Year & independent  & Easy                         & Moderate                    & Hard                        & Easy                         & Moderate                     & Hard                        & Easy                         & Moderate                     & \multicolumn{1}{c|}{Hard}                                                 & Easy                         & Moderate                     & Hard                         \\ \midrule
OFTNet~\cite{roddick2018orthographic}          & 2018 & No           & 1.61                         & 1.32                        & 1.00                        & 7.16                         & 5.69                         & 4.61                        & 4.07                         & 3.27                         & \multicolumn{1}{c|}{3.29}                                                 & 11.06                        & 8.79                         & 8.91                         \\
MonoGRNet~\cite{qin2019monogrnet}              & 2018 & No           & 9.61                         & 5.74                        & 4.25                        & 18.19                        & 11.17                        & 8.73                        & 13.88                        & 10.19                        & \multicolumn{1}{c|}{7.62}                                                 & 19.72                        & 12.81                        & 10.15                        \\
ROI-10D~\cite{manhardt2019roi}                 & 2019 & No           & 4.32                         & 2.02                        & 1.46                        & 9.78                         & 4.91                         & 3.74                        & 9.61                         & 6.63                         & \multicolumn{1}{c|}{6.29}                                                 & 14.50                        & 9.91                         & 8.73                         \\
MonoDIS~\cite{simonelli2019disentangling}      & 2019 & No           & 10.37                        & 7.94                        & 6.40                        & 17.23                        & 13.19                        & 11.12                       & 18.05                        & 14.98                        & \multicolumn{1}{c|}{13.42}                                                & 24.26                        & 18.43                        & 16.95                        \\
SS3D~\cite{jorgensen2019monocular}             & 2019 & No           & 10.78                        & 7.68                        & 6.51                        & 16.33                        & 11.52                        & 9.93                        & 14.52                        & 13.15                        & \multicolumn{1}{c|}{11.85}                                                & -                            & -                            & -                            \\
M3D-RPN~\cite{brazil2019m3d}                   & 2019 & No           & 14.76                        & 9.71                        & 7.42                        & 21.02                        & 13.67                        & 10.23                       & 20.40                        & 16.48                        & \multicolumn{1}{c|}{13.34}                                                & \textbf{26.86}               & 21.15                        & 17.14                        \\
Shift R-CNN~\cite{naiden2019shift}             & 2019 & No           & 8.13                         & 5.22                        & 4.78                        & 13.32                        & 8.49                         & 6.40                        & 13.84                        & 11.29                        & \multicolumn{1}{c|}{11.08}                                                & 18.61                        & 14.71                        & 13.57                        \\
MoVi-3D~\cite{simonelli2019single}             & 2020 & No           & 15.19                        & 10.90                       & 9.26                        & 22.76                        & 17.03                        & 14.85                       & 14.28                        & 11.13                        & \multicolumn{1}{c|}{9.68}                                                 & 22.36                        & 17.87                        & 15.73                        \\
SMOKE~\cite{liu2020smoke}                      & 2020 & No           & 14.03                        & 9.76                        & 7.84                        & 20.83                        & 14.49                        & 12.75                       & 14.76                        & 12.85                        & \multicolumn{1}{c|}{11.50}                                                & 19.99                        & 15.61                        & 15.28                        \\
RTM3D~\cite{li2020rtm3d}                       & 2020 & No           & 14.41                        & 10.34                       & 8.77                        & 19.17                        & 14.20                        & 11.99                       & 20.77                        & 16.86                        & \multicolumn{1}{c|}{16.63}                                                & 25.56                        & \textbf{22.12}               & \textbf{20.91}               \\
MonoPair~\cite{chen2020monopair}               & 2020 & No           & 13.04                        & 9.99                        & 8.65                        & 19.28                        & 14.83                        & 12.89                       & -                            & -                            & \multicolumn{1}{c|}{-}                                                    & -                            & -                            & -                            \\
MonoDLE~\cite{ma2021monodle}                    & 2021 & No           & 17.23                        & 12.26                       & 10.29                       & 24.79                        & 18.89                        & 16.00                       & -                            & -                            & \multicolumn{1}{c|}{-}                                                    & -                            & -                            & -                            \\
MonoFlex~\cite{zhang2021objdiff}               & 2021 & No           & \textbf{19.94}               & \textbf{13.89}              & \textbf{12.07}              & \textbf{28.23}               & \textbf{19.75}               & \textbf{16.89}              & \textbf{28.17}               & \textbf{21.92}               & \multicolumn{1}{c|}{\textbf{19.07}}                                       & -                            & -                            & -                            \\ \midrule
DeepBox3D~\cite{mousavian20173d}               & 2017 & \textbf{Yes} & 5.85                         & 4.10                        & 3.84                        & 9.99                         & 7.71                         & 5.30                        & 5.49                         & 3.96                         & \multicolumn{1}{c|}{2.92}                                                 & 9.33                         & 6.71                         & 5.11                         \\
Barabanau et al.~\cite{barabanau2019monocular} & 2019 & \textbf{Yes} & -                            & -                           & -                           & -                            & -                            & -                           & 13.96                        & 7.37                         & \multicolumn{1}{c|}{4.54}                                                 & -                            & -                            & -                            \\
Linear System~\cite{naiden2019shift}           & 2019 & \textbf{Yes} & 6.80                         & 4.14                        & 3.50                        & 11.75                        & 8.34                         & 6.80                        & 7.24                         & 5.98                         & \multicolumn{1}{c|}{5.54}                                                 & 14.74                        & 12.48                        & 11.22                        \\ \midrule
\rowcolor[HTML]{EFEFEF} 
Ours (8RP)                                     & 2020 & \textbf{Yes} & {\color[HTML]{3531FF} 15.82} & {\color[HTML]{3531FF} 7.94} & {\color[HTML]{3531FF} 6.68} & {\color[HTML]{3531FF} 22.28} & {\color[HTML]{3531FF} 11.64} & {\color[HTML]{3531FF} 9.95} & {\color[HTML]{3531FF} 19.92} & {\color[HTML]{3531FF} 15.75} & \multicolumn{1}{c|}{\cellcolor[HTML]{EFEFEF}{\color[HTML]{3531FF} 13.46}} & {\color[HTML]{3531FF} 25.32} & {\color[HTML]{3531FF} 18.09} & {\color[HTML]{3531FF} 17.42} \\
\rowcolor[HTML]{EFEFEF} 
Ours (2RP)                                     & 2020 & \textbf{Yes} & 15.21                        & 7.66                        & 6.24                        & 20.42                        & 10.96                        & 9.23                        & 18.68                        & 12.35                        & \multicolumn{1}{c|}{\cellcolor[HTML]{EFEFEF}11.54}                        & 25.06                        & 16.88                        & 13.58                        \\ \bottomrule

\end{tabular}%
}
\caption{Results on \textit{Car} (0.7 IoU) for both the KITTI3D test and evaluation set. \textbf{Bold} refers to best performance across all methods, {\color[HTML]{3531FF} blue} refers to best performance across camera independent methods. Note that the $R_{40}$ variant of AP is used for the test set, while the $R_{11}$ variant is used for the evaluation set for comparison purposes. Our method outperforms the other camera independent methods.}
\label{tab:results-relatedwork}
\end{table*}

\section{Conclusion}
\label{section:conclusion}

    In this work we present a novel %instance segmentation based 
    approach for monocular 3D object detection.
    Leveraging camera independent 2D reference points enables us to handle different camera types in one multi-task CNN. We use a proposal free instance based method which eliminates the need for NMS.
    This work is the first to take advantage of its camera independence by combining large scale public datasets for better generalisation across viewpoints.
    We show the benefits
    %strength of our method 
    by comparing to other methods in a cross-dataset context.
    % We predict object dimensions, 3D reference points projected in 2D image space and viewing angle values for every pixel.
    We outperform other camera independent methods on the challenging KITTI3D benchmark and show qualitatively how our model can cope with fisheye images even without fisheye training data.
    We briefly discuss  
    %investigate the 
    trade-offs between accuracy and speed, which is important for practical applications.
    Further we will release KITTI3D instance segmentation annotations.
    
    We believe camera independence is key in exploiting multiple datasets, 
    % We believe that camera independence is an important aspect to be investigated when anticipating the exploitation of different datasets and extending the use of your cnn-model to different camera types, even those that are not included yet in your training set.
    and hope to see more research in this direction. For future work, there are multiple interesting research topics. 
    % We hope to see more research in this direction, and as for future work, there are multiple interesting research topics.
    %
    First, the impact of the imbalance of annotation dimensions between the different datasets should be investigated. 
    Also tracking methods can be explored which leverage the distributions of RPs provided by our method. 
    Finally, clever practices from recent works (e.g. ensembles~\cite{zhang2021objdiff}, discarding distant samples~\cite{ma2021monodle} and pairwise constraints~\cite{chen2020monopair}),
    can be combined with our method to close the gap between camera dependent and independent methods.

{\small
\bibliographystyle{ieeetr}
\bibliography{References/mono3d}

\begin{thebibliography}{10}

\bibitem{Geiger2012CVPR}
A.~Geiger, P.~Lenz, and R.~Urtasun, ``Are we ready for autonomous driving? the
  kitti vision benchmark suite,'' in {\em Proceedings of the IEEE conference on
  computer vision and pattern recognition}, 2012.

\bibitem{cordts2016cityscapes}
M.~Cordts, M.~Omran, S.~Ramos, T.~Rehfeld, M.~Enzweiler, R.~Benenson,
  U.~Franke, S.~Roth, and B.~Schiele, ``The cityscapes dataset for semantic
  urban scene understanding,'' in {\em Proceedings of the IEEE conference on
  computer vision and pattern recognition}, 2016.

\bibitem{caesar2019nuscenes}
H.~Caesar, V.~Bankiti, A.~H. Lang, S.~Vora, V.~E. Liong, Q.~Xu, A.~Krishnan,
  Y.~Pan, G.~Baldan, and O.~Beijbom, ``nuscenes: A multimodal dataset for
  autonomous driving,'' in {\em Proceedings of the IEEE Conference on Computer
  Vision and Pattern Recognition}, pp.~11621--11631, 2020.

\bibitem{sun2019scalability}
P.~Sun, H.~Kretzschmar, X.~Dotiwalla, A.~Chouard, V.~Patnaik, P.~Tsui, J.~Guo,
  Y.~Zhou, Y.~Chai, B.~Caine, {\em et~al.}, ``Scalability in perception for
  autonomous driving: Waymo open dataset,'' in {\em Proceedings of the IEEE
  Conference on Computer Vision and Pattern Recognition}, pp.~2446--2454, 2020.

\bibitem{lyft2019}
R.~Kesten, M.~Usman, J.~Houston, T.~Pandya, K.~Nadhamuni, A.~Ferreira, M.~Yuan,
  B.~Low, A.~Jain, P.~Ondruska, S.~Omari, S.~Shah, A.~Kulkarni, A.~Kazakova,
  C.~Tao, L.~Platinsky, W.~Jiang, and V.~Shet, ``Lyft level 5 av dataset
  2019.'' url{https://level5.lyft.com/dataset/}, 2019.

\bibitem{sturm2011}
P.~Sturm, S.~Ramalingam, J.-P. Tardif, S.~Gasparini, and J.~Barreto, ``Camera
  models and fundamental concepts used in geometric computer vision,'' {\em
  Foundations and Trends® in Computer Graphics and Vision}, vol.~6, no.~1–2,
  pp.~1--183, 2011.

\bibitem{ye2020sarpnet}
Y.~Ye, H.~Chen, C.~Zhang, X.~Hao, and Z.~Zhang, ``Sarpnet: Shape attention
  regional proposal network for lidar-based 3d object detection,'' {\em
  Neurocomputing}, vol.~379, pp.~53--63, 2020.

\bibitem{hestructure}
C.~He, H.~Zeng, J.~Huang, X.-S. Hua, and L.~Zhang, ``Structure aware
  single-stage 3d object detection from point cloud,'' in {\em Proceedings of
  the IEEE Conference on Computer Vision and Pattern Recognition},
  pp.~11873--11882, 2020.

\bibitem{chen2020dsgn}
Y.~Chen, S.~Liu, X.~Shen, and J.~Jia, ``Dsgn: Deep stereo geometry network for
  3d object detection,'' in {\em Proceedings of the IEEE Conference on Computer
  Vision and Pattern Recognition}, pp.~12536--12545, 2020.

\bibitem{xu2020zoomnet}
Z.~Xu, W.~Zhang, X.~Ye, X.~Tan, W.~Yang, S.~Wen, E.~Ding, A.~Meng, and
  L.~Huang, ``Zoomnet: Part-aware adaptive zooming neural network for 3d object
  detection.,'' in {\em AAAI}, pp.~12557--12564, 2020.

\bibitem{wang2019pseudo}
Y.~Wang, W.-L. Chao, D.~Garg, B.~Hariharan, M.~Campbell, and K.~Q. Weinberger,
  ``Pseudo-lidar from visual depth estimation: Bridging the gap in 3d object
  detection for autonomous driving,'' in {\em Proceedings of the IEEE
  Conference on Computer Vision and Pattern Recognition}, pp.~8445--8453, 2019.

\bibitem{you2019pseudo}
Y.~You, Y.~Wang, W.-L. Chao, D.~Garg, G.~Pleiss, B.~Hariharan, M.~Campbell, and
  K.~Q. Weinberger, ``Pseudo-lidar++: Accurate depth for 3d object detection in
  autonomous driving,'' in {\em International Conference on Learning
  Representations}, 2019.

\bibitem{mousavian20173d}
A.~Mousavian, D.~Anguelov, J.~Flynn, and J.~Kosecka, ``3d bounding box
  estimation using deep learning and geometry,'' in {\em Proceedings of the
  IEEE Conference on Computer Vision and Pattern Recognition}, pp.~7074--7082,
  2017.

\bibitem{liu2019deep}
L.~Liu, J.~Lu, C.~Xu, Q.~Tian, and J.~Zhou, ``Deep fitting degree scoring
  network for monocular 3d object detection,'' in {\em Proceedings of the IEEE
  Conference on Computer Vision and Pattern Recognition}, pp.~1057--1066, 2019.

\bibitem{naiden2019shift}
A.~Naiden, V.~Paunescu, G.~Kim, B.~Jeon, and M.~Leordeanu, ``Shift r-cnn: Deep
  monocular 3d object detection with closed-form geometric constraints,'' in
  {\em 2019 IEEE International Conference on Image Processing}, pp.~61--65,
  IEEE, 2019.

\bibitem{min2019multi}
H.~Min~Choi, H.~Kang, and Y.~Hyun, ``Multi-view reprojection architecture for
  orientation estimation,'' in {\em Proceedings of the IEEE International
  Conference on Computer Vision Workshops}, pp.~0--0, 2019.

\bibitem{ku2019monocular}
J.~Ku, A.~D. Pon, and S.~L. Waslander, ``Monocular 3d object detection
  leveraging accurate proposals and shape reconstruction,'' in {\em Proceedings
  of the IEEE Conference on Computer Vision and Pattern Recognition},
  pp.~11867--11876, 2019.

\bibitem{chen2016monocular}
X.~Chen, K.~Kundu, Z.~Zhang, H.~Ma, S.~Fidler, and R.~Urtasun, ``Monocular 3d
  object detection for autonomous driving,'' in {\em Proceedings of the IEEE
  Conference on Computer Vision and Pattern Recognition}, pp.~2147--2156, 2016.

\bibitem{roddick2018orthographic}
T.~Roddick, A.~Kendall, and R.~Cipolla, ``Orthographic feature transform for
  monocular 3d object detection,'' {\em British Machine Vision Conference},
  2019.

\bibitem{kim2019deep}
Y.~Kim and D.~Kum, ``Deep learning based vehicle position and orientation
  estimation via inverse perspective mapping image,'' in {\em 2019 IEEE
  Intelligent Vehicles Symposium}, pp.~317--323, IEEE, 2019.

\bibitem{srivastava2019learning}
S.~Srivastava, F.~Jurie, and G.~Sharma, ``Learning 2d to 3d lifting for object
  detection in 3d for autonomous vehicles,'' in {\em 2019 IEEE/RSJ
  International Conference on Intelligent Robots and Systems}, pp.~4504--4511,
  IEEE, 2019.

\bibitem{reading2021cadnn}
C.~Reading, A.~Harakeh, J.~Chae, and S.~L. Waslander, ``Categorical depth
  distribution network for monocular 3d object detection,'' in {\em Proceedings
  of the IEEE Conference on Computer Vision and Pattern Recognition},
  pp.~8555--8564, 2021.

\bibitem{ZhuJ2019objdist}
J.~Zhu and Y.~Fang, ``Learning object-specific distance from a monocular
  image,'' in {\em Proceedings of the IEEE International Conference on Computer
  Vision}, pp.~3839--3848, 2019.

\bibitem{zhou2019objects}
X.~Zhou, D.~Wang, and P.~Kr{\"{a}}henb{\"{u}}hl, ``Objects as points,'' {\em
  CoRR}, vol.~abs/1904.07850, 2019.

\bibitem{qin2019monogrnet}
Z.~Qin, J.~Wang, and Y.~Lu, ``Monogrnet: A geometric reasoning network for
  monocular 3d object localization,'' in {\em Proceedings of the AAAI
  Conference on Artificial Intelligence}, vol.~33, pp.~8851--8858, 2019.

\bibitem{manhardt2019roi}
F.~Manhardt, W.~Kehl, and A.~Gaidon, ``Roi-10d: Monocular lifting of 2d
  detection to 6d pose and metric shape,'' in {\em Proceedings of the IEEE
  Conference on Computer Vision and Pattern Recognition}, pp.~2069--2078, 2019.

\bibitem{simonelli2019disentangling}
A.~Simonelli, S.~R. Bulo, L.~Porzi, M.~L{\'o}pez-Antequera, and
  P.~Kontschieder, ``Disentangling monocular 3d object detection,'' in {\em
  Proceedings of the IEEE International Conference on Computer Vision},
  pp.~1991--1999, 2019.

\bibitem{jorgensen2019monocular}
E.~J{\"o}rgensen, C.~Zach, and F.~Kahl, ``Monocular 3d object detection and box
  fitting trained end-to-end using intersection-over-union loss,'' {\em arXiv
  preprint arXiv:1906.08070}, 2019.

\bibitem{brazil2019m3d}
G.~Brazil and X.~Liu, ``M3d-rpn: Monocular 3d region proposal network for
  object detection,'' in {\em Proceedings of the IEEE International Conference
  on Computer Vision}, pp.~9287--9296, 2019.

\bibitem{simonelli2019single}
A.~Simonelli, S.~R. Bulo, L.~Porzi, E.~Ricci, and P.~Kontschieder, ``Towards
  generalization across depth for monocular 3d object detection,'' in {\em 16th
  European Conference on Computer Vision--ECCV 2020}, pp.~767--782, 2020.

\bibitem{luo2021m3dssd}
S.~Luo, H.~Dai, L.~Shao, and Y.~Ding, ``{M3DSSD:} monocular 3d single stage
  object detector,'' in {\em Proceedings of the IEEE Conference on Computer
  Vision and Pattern Recognition}, 2021.

\bibitem{ma2021monodle}
X.~Ma, Y.~Zhang, D.~Xu, D.~Zhou, S.~Yi, H.~Li, and W.~Ouyang, ``Delving into
  localization errors for monocular 3d object detection,'' in {\em Proceedings
  of the IEEE Conference on Computer Vision and Pattern Recognition}, June
  2021.

\bibitem{barabanau2019monocular}
I.~Barabanau, A.~Artemov, E.~Burnaev, and V.~Murashkin, ``Monocular 3d object
  detection via geometric reasoning on keypoints,'' {\em arXiv preprint
  arXiv:1905.05618}, 2019.

\bibitem{chabot2017deep}
F.~Chabot, M.~Chaouch, J.~Rabarisoa, C.~Teuli{\`e}re, and T.~Chateau, ``Deep
  manta: A coarse-to-fine many-task network for joint 2d and 3d vehicle
  analysis from monocular image,'' in {\em Proceedings of the IEEE Conference
  on Computer Vision and Pattern Recognition}, pp.~2040--2049, 2017.

\bibitem{kundu20183d}
A.~Kundu, Y.~Li, and J.~M. Rehg, ``3d-rcnn: Instance-level 3d object
  reconstruction via render-and-compare,'' in {\em Proceedings of the IEEE
  Conference on Computer Vision and Pattern Recognition}, pp.~3559--3568, 2018.

\bibitem{he2019mono3d++}
T.~He and S.~Soatto, ``Mono3d++: Monocular 3d vehicle detection with two-scale
  3d hypotheses and task priors,'' in {\em Proceedings of the AAAI Conference
  on Artificial Intelligence}, vol.~33, pp.~8409--8416, 2019.

\bibitem{rangesh2018ground}
A.~Rangesh and M.~M. Trivedi, ``Ground plane polling for 6dof pose estimation
  of objects on the road,'' {\em IEEE Transactions on Intelligent Vehicles},
  2020.

\bibitem{li2020rtm3d}
P.~Li, H.~Zhao, P.~Liu, and F.~Cao, ``Rtm3d: Real-time monocular 3d detection
  from object keypoints for autonomous driving,'' in {\em 16th European
  Conference on Computer Vision--ECCV 2020}, pp.~644--660, 2020.

\bibitem{cai2020monocular}
Y.~Cai, B.~Li, Z.~Jiao, H.~Li, X.~Zeng, and X.~Wang, ``Monocular 3d object
  detection with decoupled structured polygon estimation and height-guided
  depth estimation,'' in {\em Proceedings of the AAAI Conference on Artificial
  Intelligence}, vol.~34, pp.~10478--10485, 2020.

\bibitem{suwajanakorn2018discovery}
S.~Suwajanakorn, N.~Snavely, J.~J. Tompson, and M.~Norouzi, ``Discovery of
  latent 3d keypoints via end-to-end geometric reasoning,'' in {\em Advances in
  Neural Information Processing Systems}, pp.~2059--2070, 2018.

\bibitem{liu2020smoke}
Z.~Liu, Z.~Wu, and R.~T{\'o}th, ``Smoke: Single-stage monocular 3d object
  detection via keypoint estimation,'' in {\em Proceedings of the IEEE
  Conference on Computer Vision and Pattern Recognition Workshops},
  pp.~996--997, 2020.

\bibitem{chen2020monopair}
Y.~Chen, L.~Tai, K.~Sun, and M.~Li, ``Monopair: Monocular 3d object detection
  using pairwise spatial relationships,'' in {\em Proceedings of the IEEE
  Conference on Computer Vision and Pattern Recognition}, pp.~12093--12102,
  2020.

\bibitem{pavlakos20176}
G.~Pavlakos, X.~Zhou, A.~Chan, K.~G. Derpanis, and K.~Daniilidis, ``6-dof
  object pose from semantic keypoints,'' in {\em 2017 IEEE International
  Conference on Robotics and Automation}, pp.~2011--2018, IEEE, 2017.

\bibitem{Yang2021LiteFPNFK}
L.~Yang, X.~Zhang, L.~Wang, M.~Zhu, and J.~Li, ``Lite-fpn for keypoint-based
  monocular 3d object detection,'' {\em ArXiv}, vol.~abs/2105.00268, 2021.

\bibitem{xu2018multi}
B.~Xu and Z.~Chen, ``Multi-level fusion based 3d object detection from
  monocular images,'' in {\em Proceedings of the IEEE Conference on Computer
  Vision and Pattern Recognition}, pp.~2345--2353, 2018.

\bibitem{ma2019accurate}
X.~Ma, Z.~Wang, H.~Li, P.~Zhang, W.~Ouyang, and X.~Fan, ``Accurate monocular 3d
  object detection via color-embedded 3d reconstruction for autonomous
  driving,'' in {\em Proceedings of the IEEE International Conference on
  Computer Vision}, pp.~6851--6860, 2019.

\bibitem{weng2019monocular}
X.~Weng and K.~Kitani, ``Monocular 3d object detection with pseudo-lidar point
  cloud,'' in {\em 2019 IEEE International Conference on Computer Vision
  Workshop}, pp.~857--866, IEEE Computer Society, 2019.

\bibitem{vianney2019refinedmpl}
J.~M.~U. Vianney, S.~Aich, and B.~Liu, ``Refinedmpl: Refined monocular
  pseudolidar for 3d object detection in autonomous driving,'' {\em arXiv
  preprint arXiv:1911.09712}, 2019.

\bibitem{qian2020end}
R.~Qian, D.~Garg, Y.~Wang, Y.~You, S.~Belongie, B.~Hariharan, M.~Campbell,
  K.~Q. Weinberger, and W.-L. Chao, ``End-to-end pseudo-lidar for image-based
  3d object detection,'' in {\em Proceedings of the IEEE Conference on Computer
  Vision and Pattern Recognition}, pp.~5881--5890, 2020.

\bibitem{song2015joint}
S.~Song and M.~Chandraker, ``Joint sfm and detection cues for monocular 3d
  localization in road scenes,'' in {\em Proceedings of the IEEE Conference on
  Computer Vision and Pattern Recognition}, pp.~3734--3742, 2015.

\bibitem{Li2021depthmessprop}
L.~Wang, L.~Du, X.~Ye, Y.~Fu, G.~Guo, X.~Xue, J.~Feng, and L.~Zhang,
  ``Depth-conditioned dynamic message propagation for monocular 3d object
  detection,'' in {\em Proceedings of the IEEE Conference on Computer Vision
  and Pattern Recognition}, 2021.

\bibitem{ding2020learning}
M.~Ding, Y.~Huo, H.~Yi, Z.~Wang, J.~Shi, Z.~Lu, and P.~Luo, ``Learning
  depth-guided convolutions for monocular 3d object detection,'' in {\em
  Proceedings of the IEEE Conference on Computer Vision and Pattern Recognition
  Workshops}, pp.~1000--1001, 2020.

\bibitem{Liu2021GroundAwareM3}
Y.~Liu, Y.~Yixuan, and M.~Liu, ``Ground-aware monocular 3d object detection for
  autonomous driving,'' {\em IEEE Robotics and Automation Letters}, vol.~6,
  pp.~919--926, 2021.

\bibitem{zhou2021monoEF}
Y.~Zhou, Y.~He, H.~Zhu, C.~Wang, H.~Li, and Q.~Jiang, ``Monocular 3d object
  detection: An extrinsic parameter free approach,'' in {\em Proceedings of the
  IEEE Conference on Computer Vision and Pattern Recognition}, 2021.

\bibitem{Chen2021MonoRUnM3}
H.~Chen, Y.~Huang, W.~Tian, Z.~Gao, and L.~Xiong, ``Monorun: Monocular 3d
  object detection by reconstruction and uncertainty propagation,'' in {\em
  Proceedings of the IEEE Conference on Computer Vision and Pattern
  Recognition}, pp.~10379--10388, 2021.

\bibitem{zhang2021objdiff}
Z.~Yunpeng, L.~Jiwen, and Z.~Jie, ``Objects are different: Flexible monocular
  3d object detection,'' in {\em Proceedings of the IEEE Conference on Computer
  Vision and Pattern Recognition}, 2021.

\bibitem{xiang2018posecnn}
Y.~Xiang, T.~Schmidt, V.~Narayanan, and D.~Fox, ``Posecnn: A convolutional
  neural network for 6d object pose estimation in cluttered scenes,'' in {\em
  Proceedings of Robotics: Science and Systems}, (Pittsburgh, Pennsylvania),
  June 2018.

\bibitem{capellen2019convposecnn}
C.~Capellen., M.~Schwarz., and S.~Behnke., ``Convposecnn: Dense convolutional
  6d object pose estimation,'' in {\em Proceedings of the 15th International
  Joint Conference on Computer Vision, Imaging and Computer Graphics Theory and
  Applications}, pp.~162--172, 2020.

\bibitem{do2018deep}
T.~Do, T.~Pham, M.~Cai, and I.~Reid, ``Lienet: Real-time monocular object
  instance 6d pose estimation,'' in {\em British Machine Vision Conference
  2018, {BMVC} 2018, Newcastle, UK, September 3-6, 2018}, p.~2, 2018.

\bibitem{jafari2018ipose}
O.~H. Jafari, S.~K. Mustikovela, K.~Pertsch, E.~Brachmann, and C.~Rother,
  ``ipose: instance-aware 6d pose estimation of partly occluded objects,'' in
  {\em Asian Conference on Computer Vision}, pp.~477--492, Springer, 2018.

\bibitem{li2019cdpn}
Z.~Li, G.~Wang, and X.~Ji, ``Cdpn: Coordinates-based disentangled pose network
  for real-time rgb-based 6-dof object pose estimation,'' in {\em Proceedings
  of the IEEE International Conference on Computer Vision}, pp.~7678--7687,
  2019.

\bibitem{zakharov2019dpod}
S.~Zakharov, I.~Shugurov, and S.~Ilic, ``Dpod: 6d pose object detector and
  refiner,'' in {\em Proceedings of the IEEE International Conference on
  Computer Vision}, pp.~1941--1950, 2019.

\bibitem{wang2019normalized}
H.~Wang, S.~Sridhar, J.~Huang, J.~Valentin, S.~Song, and L.~J. Guibas,
  ``Normalized object coordinate space for category-level 6d object pose and
  size estimation,'' in {\em Proceedings of the IEEE Conference on Computer
  Vision and Pattern Recognition}, pp.~2642--2651, 2019.

\bibitem{Fan2021DeepLO}
Z.~Fan, Y.~Zhu, Y.~He, Q.~Sun, H.~Liu, and J.~He, ``Deep learning on monocular
  object pose detection and tracking: A comprehensive overview,'' {\em ArXiv},
  vol.~abs/2105.14291, 2021.

\bibitem{peng2019pvnet}
S.~Peng, Y.~Liu, Q.~Huang, X.~Zhou, and H.~Bao, ``Pvnet: Pixel-wise voting
  network for 6dof pose estimation,'' in {\em Proceedings of the IEEE
  Conference on Computer Vision and Pattern Recognition}, pp.~4561--4570, 2019.

\bibitem{hu2019segmentation}
Y.~Hu, J.~Hugonot, P.~Fua, and M.~Salzmann, ``Segmentation-driven 6d object
  pose estimation,'' in {\em Proceedings of the IEEE Conference on Computer
  Vision and Pattern Recognition}, pp.~3385--3394, 2019.

\bibitem{romera2017erfnet}
E.~Romera, J.~M. Alvarez, L.~M. Bergasa, and R.~Arroyo, ``Erfnet: Efficient
  residual factorized convnet for real-time semantic segmentation,'' {\em IEEE
  Transactions on Intelligent Transportation Systems}, vol.~19, no.~1,
  pp.~263--272, 2017.

\bibitem{he2016deep}
K.~He, X.~Zhang, S.~Ren, and J.~Sun, ``Deep residual learning for image
  recognition,'' in {\em Proceedings of the IEEE conference on computer vision
  and pattern recognition}, pp.~770--778, 2016.

\bibitem{torchvisionmodels}
{PyTorch}, ``{Documentation: torchvision.models}.''
  \url{https://pytorch.org/docs/stable/torchvision/models.html}.
\newblock Accessed: 2020-11.

\bibitem{neven2019instance}
D.~Neven, B.~D. Brabandere, M.~Proesmans, and L.~V. Gool, ``Instance
  segmentation by jointly optimizing spatial embeddings and clustering
  bandwidth,'' in {\em Proceedings of the IEEE Conference on Computer Vision
  and Pattern Recognition}, pp.~8837--8845, 2019.

\bibitem{githubSMOKE}
Z.~Liu, Z.~Wu, and R.~T\'oth, ``{SMOKE: Single-Stage Monocular 3D Object
  Detection via Keypoint Estimation}.'' \url{https://github.com/lzccccc/SMOKE},
  2020.
\newblock Accessed: 2020-09.

\bibitem{githubM3DRPN}
G.~Brazil and X.~Liu, ``{M3D-RPN: Monocular 3D Region Proposal Network for
  Object Detection}.'' \url{https://github.com/garrickbrazil/M3D-RPN}, 2019.
\newblock Accessed: 2020-09.

\bibitem{massa2018mrcnn}
F.~Massa and R.~Girshick, ``{maskrcnn-benchmark: Fast, modular reference
  implementation of Instance Segmentation and Object Detection algorithms in
  PyTorch}.'' \url{https://github.com/facebookresearch/maskrcnn-benchmark},
  2018.
\newblock Accessed: 2020-11.

\bibitem{gaidon2016virtual}
A.~Gaidon, Q.~Wang, Y.~Cabon, and E.~Vig, ``Virtual worlds as proxy for
  multi-object tracking analysis,'' in {\em Proceedings of the IEEE conference
  on computer vision and pattern recognition}, pp.~4340--4349, 2016.

\bibitem{Alhaija2018IJCV}
H.~A. Alhaija, S.~K. Mustikovela, L.~Mescheder, A.~Geiger, and C.~Rother,
  ``Augmented reality meets computer vision: Efficient data generation for
  urban driving scenes,'' {\em International Journal of Computer Vision},
  vol.~126, no.~9, pp.~961--972, 2018.

\bibitem{wang2020train}
Y.~Wang, X.~Chen, Y.~You, L.~E. Li, B.~Hariharan, M.~Campbell, K.~Q.
  Weinberger, and W.-L. Chao, ``Train in germany, test in the usa: Making 3d
  object detectors generalize,'' in {\em Proceedings of the IEEE Conference on
  Computer Vision and Pattern Recognition}, pp.~11713--11723, 2020.

\bibitem{tensorrt}
{NVIDIA Corporation}, ``{NVIDIA TensorRT: Programmable Inference
  Accelerator}.'' \url{https://developer.nvidia.com/tensorrt}.
\newblock Accessed: 2020-11.

\bibitem{yogamani2019woodscape}
S.~Yogamani, C.~Hughes, J.~Horgan, G.~Sistu, P.~Varley, D.~O'Dea,
  M.~Uric{\'a}r, S.~Milz, M.~Simon, K.~Amende, {\em et~al.}, ``Woodscape: A
  multi-task, multi-camera fisheye dataset for autonomous driving,'' in {\em
  Proceedings of the IEEE International Conference on Computer Vision},
  pp.~9308--9318, 2019.

\end{thebibliography}
%\bibliography{References/added_for_iccv}
}

\end{document}